\documentclass[sigconf]{acmart}
\AtBeginDocument{%
  }
\usepackage{makecell}
\usepackage{booktabs}
\usepackage{balance}
\usepackage{hyperref}
\hypersetup{
    colorlinks=true,  
    linkcolor=blue,   
    urlcolor=cyan,    
    citecolor=green,  
}

\copyrightyear{2025}
\acmYear{2025}
\setcopyright{acmlicensed}\acmConference[MM '25]{Proceedings of the 33rd
ACM International Conference on Multimedia}{October 27--31, 2025}{Dublin,
Ireland}
\acmBooktitle{Proceedings of the 33rd ACM International Conference on
Multimedia (MM '25), October 27--31, 2025, Dublin, Ireland}
\acmDOI{10.1145/3746027.3754704}
\acmISBN{979-8-4007-2035-2/2025/10}

\begin{document}

\title{TiP4GEN: Text to Immersive Panorama 4D Scene Generation}

\author{Ke Xing}
\authornote{Both authors contributed equally to this research.}
\orcid{0009-0004-2626-9072}
\affiliation{%
  \institution{\normalsize Institute of Information Science, Beijing Jiaotong University}
  \vspace{-1pt}\institution{\normalsize Visual Intelligence + X International Joint Laboratory}
  \city{Beijing}
  \country{China}
}

\author{Hanwen Liang}
\authornotemark[1]
\orcid{0000-0002-9892-752X}
\affiliation{%
  \institution{\normalsize University of Toronto}
  \city{Toronto}
  \country{Canada}
}

\author{Dejia Xu}
\orcid{0000-0001-8474-3095}
\affiliation{%
  \institution{\normalsize The University of Texas at Austin}
  \city{Austin}
  \country{USA}
}

\author{Yuyang Yin}
\orcid{0009-0005-7443-6194}
\affiliation{%
  \institution{\normalsize Institute of Information Science, Beijing Jiaotong University}
  \vspace{-1pt}\institution{\normalsize Visual Intelligence + X International Joint Laboratory}
  \city{Beijing}
  \country{China}
}

\author{Konstantinos N. Plataniotis}
\orcid{0000-0003-3647-5473}
\affiliation{%
  \institution{\normalsize University of Toronto}
  \city{Toronto}
  \country{Canada}
}

\author{Yao Zhao}
\orcid{0000-0002-8581-9554}
\affiliation{%
  \institution{\normalsize Institute of Information Science, Beijing Jiaotong University}
  \vspace{-1pt}\institution{\normalsize Visual Intelligence + X International Joint Laboratory}
  \city{Beijing}
  \country{China}
}

\author{Yunchao Wei}
\authornote{Corresponding author.}
\orcid{0000-0002-2812-8781}
\affiliation{%
  \institution{\normalsize Institute of Information Science, Beijing Jiaotong University}
  \vspace{-1pt}\institution{\normalsize Visual Intelligence + X International Joint Laboratory}
  \city{Beijing}
  \country{China}
}

\renewcommand{\shortauthors}{Ke Xing et al.}

\begin{abstract}
With the rapid advancement and widespread adoption of VR/AR technologies, there is a growing demand for the creation of high-quality, immersive dynamic scenes.
However, existing generation works predominantly concentrate on the creation of static scenes or narrow perspective-view dynamic scenes, falling short of delivering a truly 360-degree immersive experience from any viewpoint.
In this paper, we introduce \textbf{TiP4GEN}, an advanced text-to-dynamic panorama scene generation framework that enables fine-grained content control and synthesizes motion-rich, geometry-consistent panoramic 4D scenes.
TiP4GEN integrates panorama video generation and dynamic scene reconstruction to create 360-degree immersive virtual environments.
For video generation, we introduce a \textbf{Dual-branch Generation Model} consisting of a panorama branch and a perspective branch, responsible for global and local view generation, respectively.
A bidirectional cross-attention mechanism facilitates comprehensive information exchange between the branches.
For scene reconstruction, we propose a \textbf{Geometry-aligned Reconstruction Model} based on 3D Gaussian Splatting.
By aligning spatial-temporal point clouds using metric depth maps and initializing scene cameras with estimated poses, our method ensures geometric consistency and temporal coherence for the reconstructed scenes.
Extensive experiments demonstrate the effectiveness of our proposed designs and the superiority of TiP4GEN in generating visually compelling and motion-coherent dynamic panoramic scenes. Our project page is at \url{https://ke-xing.github.io/TiP4GEN/}.
\end{abstract}



\begin{teaserfigure}
  \includegraphics[width=1.0\textwidth]{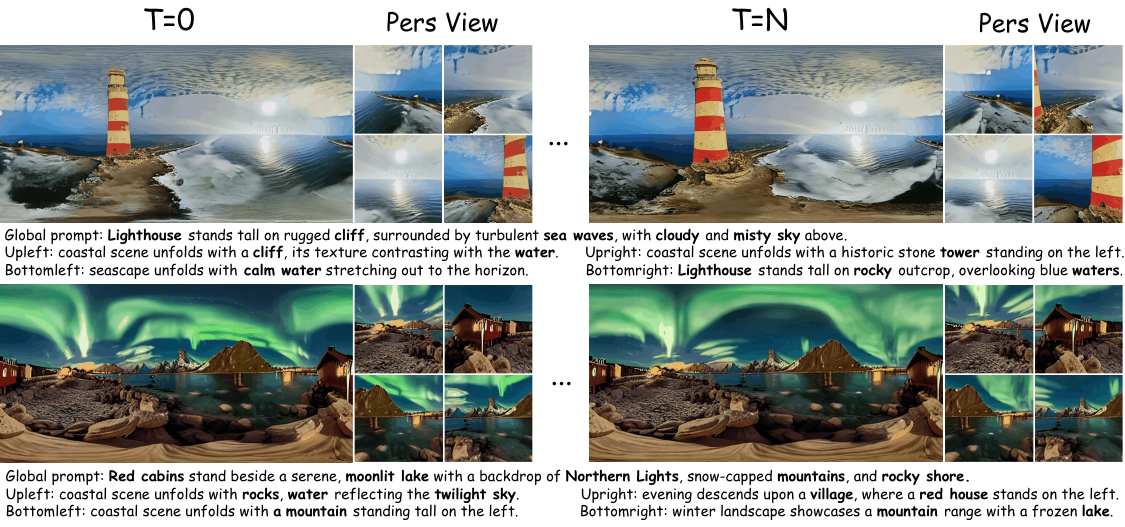}
  \caption{We introduce TiP4GEN, an advanced text-to-dynamic panorama scene generation framework capable of producing 360-degree immersive and high-fidelity 4D scenes. 
 Given a global text prompt and four local text prompts, TiP4GEN generates visually pleasant, motion-rich panoramic videos with fine-grained content control and then explicitly reconstructs panorama 4D scenes with high spatial-temporal geometry consistency.
 }
  \label{fig:teaser}
\end{teaserfigure}


\maketitle

\section{Introduction}
As VR/AR technologies advance at an accelerated pace and gain broader acceptance, immersive reality solutions are finding growing applications across industries such as gaming, entertainment, and education. This growth drives a rising demand for high-quality, immersive dynamic panorama scene generation.

In recent years, diffusion-based generative models have advanced significantly from image and video generation\cite{rombach2022high, blattmann2023stable} to 3D and 4D\cite{poole2022dreamfusion,yin20234dgen,wu2024cat4d} asset creation.
While recent work has expanded from object-centric asset generation to scene synthesis, most research has focused mainly on static 3D scenes~\cite{zhou2024dreamscene360,yang2024layerpano3d,chung2023luciddreamer,liang2024wonderland}, with relatively less attention to dynamic 4D scenes.
This is primarily due to the scarcity of large-scale 4D scene datasets.
Also, existing 4D scene generation methods~\cite{lee2024vividdream, xu2024comp4d, wang20244real,liang2024diffusion4d} are often constrained to narrow perspectives and struggle to maintain global consistency in full 360-degree views, thereby restricting their applicability to immersive scene experiences.
Recent works have made attempts to enhance immersion and achieved promising results by generating panoramic video from text~\cite{wang2024360dvd} or perspective videos~\cite{tan2024imagine360}, or by animating panoramic images~\cite{li20244k4dgen}.
However, there are several limitations with current approaches.
First, although panoramic scenes offer a much wider field of view than perspective scenes, prior works often generate simplistic scene compositions where a single object (e.g., a mountain or a lake) dominates the entire scene, lacking fine-grained control over local view variations.
These methods typically rely on a single global text prompt, enforcing uniform content distribution and leading to homogeneous, less diverse scenes.
Second, the spherical visual and motion patterns in panoramic video pose challenges for directly finetuning traditional perspective video diffusion models for panorama generation~\cite{tan2024imagine360}. 
Unlike perspective video generation models benefiting from training on web-scale datasets, the scarcity of panoramic video data makes it non-trivial to train a well-generalized model capable of producing content-diverse and motion-coherent panoramic videos.
Furthermore, in the 4D construction stage, prior work~\cite{li20244k4dgen} assumes a static camera and models per-timestamp geometry independently without explicit camera pose estimation, making it difficult to maintain temporal coherence, especially when the scene undergoes substantial movements.

To address these limitations, we propose \textbf{TiP4GEN}, an advanced text-to-dynamic panorama scene generation framework that enables fine-grained content control and generates motion-rich and geometry-consistent 4D scenes.
To support novel view synthesis and deliver immersive viewing experiences, our framework adopts a two-stage approach: first generating semantically rich panoramic videos and then reconstructing the dynamic 4D scenes.
In the video generation phase, considering the limited panoramic video datasets, we take advantage of pretrained perspective video diffusion models and propose a \textbf{Dual-branch Generation Model}.
This model comprises a panorama branch, which maintains global consistency, and a perspective branch, which enhances local detail and diversity by leveraging pretrained priors.
A bidirectional cross-attention mechanism is adopted between the two branches to facilitate information exchange and improve the whole scene coherence.
The perspective branch generates multiple views at the same time, each guided by a distinct local text prompt, enabling fine-grained control over panoramic scene composition.
In the dynamic scene reconstruction phase, we introduce \textbf{Geometry-aligned Reconstruction Model}, which employs 3D Gaussian Splatting (3DGS)\cite{kerbl3Dgaussians} sets to represent the panoramic scenes across timestamps.
Individually reconstructing the scene of each frame overlooks the temporal continuity in video sequences, resulting in inconsistencies in the reconstructed geometry across time.
Moreover, within a single panoramic frame, multiple perspective views should also share a unified geometry under the same metric space.
To address these issues and enhance spatial-temporal coherence, we introduce spatial and temporal alignment modules that leverage estimated depths~\cite{ranftl2021vision,zhang2024monst3r} to align geometry consistently across both viewpoints and timestamps.
To mitigate the inconsistency caused by camera motion, we also estimate camera poses to initialize the scene poses, followed by per-scene GS optimization.

Our contributions can be summarized as follows:
\begin{itemize}
    \item We propose \textbf{TiP4GEN}, a novel text-to-dynamic panorama scene generation framework that synthesizes immersive 360-degree, motion-rich, and geometry-consistent 4D scenes with fine-grained content control.

    \item We introduce a \textbf{Dual-branch Generation Model} that jointly leverages panoramic and perspective branches to integrate global scene consistency and local detail diversity, with a bidirectional spatial-temporal cross-attention mechanism enabling coherent content fusion.

    \item We develop a \textbf{Geometry-aligned Reconstruction Model} that reconstructs dynamic panoramic 4D scenes using Gaussian Splatting, incorporating spatial-temporal alignment and camera pose initialization to largely enhance the geometry consistency across time and views.

    \item Extensive experiments demonstrate that our method significantly outperforms existing methods in generating semantically rich, motion-coherent, and geometrically consistent dynamic 4D panoramic scenes.
\end{itemize}

\section{Related Works}

\subsection{Video Diffusion Model}
In recent years, diffusion models\cite{ho2020denoising,song2020denoising,sohl2015deep,song2019generative} have achieved remarkable success in the field of image generation\cite{podell2023sdxlimprovinglatentdiffusion,rombach2022high,saharia2022photorealistic,nichol2021glide}, demonstrating significant advancements in both generation quality\cite{rombach2022high,podell2023sdxlimprovinglatentdiffusion,saharia2022photorealistic} and efficiency\cite{karras2022elucidating,NEURIPS2022_260a14ac,song2020denoising}. This success has spurred interest in exploring the broader applications of diffusion models, particularly in video generation\cite{blattmann2023stable,deng2023mv} and the creation of 3D\cite{poole2022dreamfusion,liu2023one,shi2023mvdream,chung2023luciddreamer,liang2024wonderland} and 4D\cite{yin20234dgen,liang2024diffusion4d,li20244k4dgen,lee2024vividdream} assets.
Researchers have attempted to leverage the knowledge priors from pre-trained video diffusion models to address data insufficiency in 3D/4D generation by generating multiple views of the assets. For instance, ~\cite{voleti2024sv3d, zuo2024videomv} integrate the spatial-temporal modeling in video diffusion models to enable the generation of 3D assets. \cite{liang2024diffusion4d} fine-tunes the 3D-aware video diffusion model to enhance its capability to generate multiple views of dynamic 3D objects and enable 4D asset creation. However, the body of literature and research about 4D scene generation remains limited. Here, we leverage the robust perspective generation capability of the standard video diffusion model to create panorama videos and transform them into 4D scenes.

\subsection{Panorama Video Generation}
Recently, significant advancements have been made in the generation of immersive large-scale 3D scenes, such as LucidDreamer\cite{chung2023luciddreamer}, Wonderland\cite{liang2024wonderland}, and Wonderworld\cite{yu2024wonderworld}. However, these methods have primarily focused on static 3D environments and offered limited observable perspective angles due to the reliance on outpainting techniques. Consequently, attention has shifted towards panorama images for more comprehensive immersive experiences. Early explorations in generating panoramas using GAN-based methods were conducted, with models like OmniDreamer\cite{akimoto2022diverse} employing circular inference strategies to ensure horizontal closed-loop continuity. ImmenseGAN\cite{dastjerdi2022guided} introduced fine-tuning to enhance control over the generated output. Recently, diffusion-based approaches have emerged as dominant in panorama generation. For instance, PanoGen\cite{li2023panogen} leverages a latent diffusion model to synthesize indoor panorama images, while StitchDiffusion\cite{wang2024customizing} employs a T2I diffusion model to maintain both-ends continuity in the generated panoramas. PanFusion\cite{zhang2024taming} introduced a dual-branch design to address issues such as loop inconsistency, distorted lines, repetitive objects, and unreasonable furniture layouts. 360DVD\cite{wang2024360dvd} has pioneered the generation of panorama video. In addition to that, Dreamscene360\cite{zhou2024dreamscene360} uses 3D Gaussian Splatting\cite{kerbl3Dgaussians} to expand the dimensionality of panorama images, achieving the first successful generation of immersive panorama scenes.

\subsection{Dynamic Scene Generation}

With the advancement of generative artificial intelligence, there has been increasing interest in the creation of dynamic scenes. An ideal dynamic scene should be immersive, observable from any angle of view, with rich motion forms and amplitudes, and of high-quality refinement. Several significant studies have been conducted recently. Vividdream\cite{lee2024vividdream} generates dynamic scenes by outpainting and animating perspective views; however, its generated scene perspectives are constrained, limiting immersive observation from arbitrary views. 4K4DGen\cite{li20244k4dgen} addresses this limitation by generating dynamic scenes through the animation of panorama images, thereby resolving the issue of restricted views. Nonetheless, it encounters challenges such as a limited range of motion, monotonous types of movement, and predominantly background-only motion. Moreover, since 4K4DGen relies on panorama images as input, it lacks the capability for fine-grained control over the content of the generated scenes. In contrast, our method, which employs text-based control, can produce high-quality, motion-rich, immersive dynamic panorama scenes.

\section{Method}
In this section, we present our framework, TiP4GEN, for dynamic panorama scene generation.
We begin with a concise review of the latent diffusion model and 3D Gaussian Splatting (3DGS).
Then, we introduce the two key components of TiP4GEN: dual-branch panorama video generation and explicit 4D panorama scene reconstruction.
The dual-branch architecture comprises a panorama branch to ensure global scene coherence and a perspective branch to enable fine-grained generation control and improve content diversity.
A bidirectional cross-attention mechanism enables effective interaction between the branches, ensuring consistent and coherent scene generation in two branches.
For 4D scene reconstruction, we apply spatial and temporal alignment to the scene’s point clouds to enforce geometry consistency across space and time. Camera pose estimation is also used to initialize scene poses, mitigating inconsistencies caused by camera motion.
\subsection{Preliminaries}

\textbf{Latent Diffusion Model.}
Diffusion models define a forward process that gradually adds noise to an input signal and learn to reverse this process by denoising.
Modern video diffusion models mostly operate in latent space for efficiency~\cite{hong2022cogvideo, blattmann2023stable,liu2024sora,he2022latent,blattmann2023align}. 
Given a source video $x\in R^{T\times 3\times H\times W}$ with spatial dimension $H\times W$ and $T$ frames, a pretrained VAE encoder $\mathcal{E}$ first embeds the video to a latent representation $z$.
Random noise $\epsilon \sim \mathcal{N}(0,I)$ is added to $z$ to get the noisy latent, which is then input to a denoising network $\epsilon _{\theta} $ (parameterized by $\theta$) to perform the denoising process.
The training objective is as follows:
\begin{equation}
    \label {eq:ldm_loss} \mathcal {L} = \mathbb {E}_{{\mathcal {E}}({x}),{t},{\epsilon },{y}} \left [ || {\epsilon } - {\epsilon }_{\theta }({z}_{t},{t},y) ||_2 \right ],
\end{equation}
where $y$ denotes conditional inputs (e.g., text or image prompt), $z_t$ denotes the noisy latent with noise magnitude parameterized by diffusion timestamp $t$.
During inference, the denoised clean latent is decoded into the image space by the VAE decoder $\mathcal{D}$.

\textbf{3D Gaussian Splatting.}
3DGS represents a scene using a set of 3D Gaussians, offering an expressive and differentiable rendering format.
Each Gaussian is parameterized as $g \equiv (\mu, R, s, o, c)$, where $\mu \in \mathbb{R}^{3}$ denotes the center position of the 3D Gaussian, $R \in \mathbb{SO}(3)$ represents the orientation, $s \in \mathbb{R}^{3}$ specifies the scale, $o \in \mathbb{R}$ indicates the opacity, and $c \in \mathbb{R}^{3}$ signifies the color. 

\begin{figure*}[ht]
    \centering
    \includegraphics[width=\textwidth]{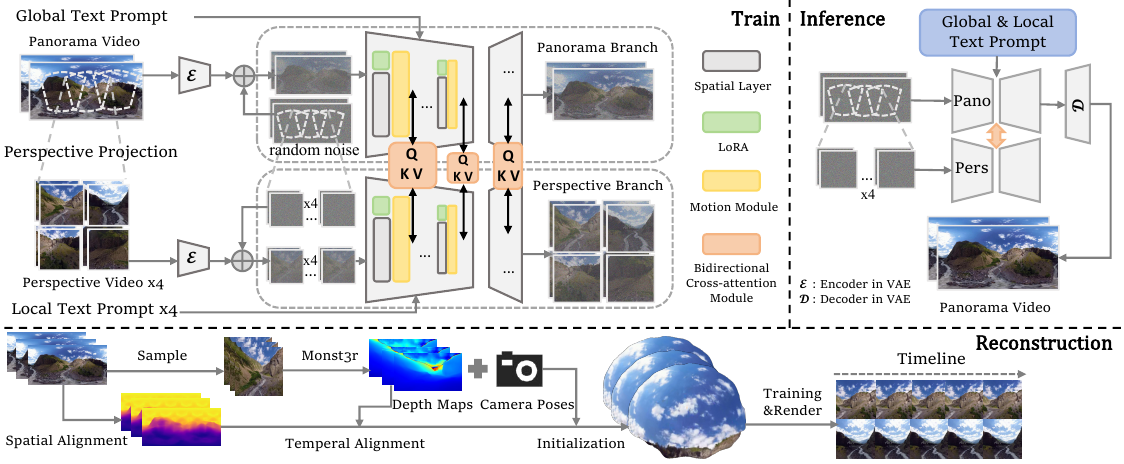}
    \caption{The framework of TiP4GEN. In the dual-branch panorama video generation model (upper), we employ a panorama branch to provide global guidance, ensuring the overall consistency of the generated content. Simultaneously, to leverage the robust knowledge priors inherent in diffusion models about perspective content generation, we incorporate a perspective branch to enhance the diversity and authenticity of the generated content. The bidirectional cross-attention module facilitates information exchange between these two branches. In the reconstruction model (lower), we utilize Monst3r for estimating camera parameters and depth maps to initialize and align the scene geometry, followed by 4D scene optimization.}
    \label{fig:framework}
\end{figure*}

\subsection{Dual-branch Panorama Video Generation}
Directly finetuning foundation video diffusion models for panoramic generation often leads to limited scene diversity and simplistic motion patterns~\cite{li20244k4dgen}, primarily due to the lack of a large-scale panoramic dataset and the domain gap between spherical projections and the linear visual and motion patterns typical of perspective videos.
However, synthesizing panorama views by generating multiple perspective views often leads to issues such as boundary inconsistency, loop closure failures\cite{tang2023mvdiffusionenablingholisticmultiview}, and repeated elements, due to a lack of global context.
In order to take advantage of the robust generative priors in the foundation video diffusion models, we propose a dual-branch architecture consisting of a panorama branch for global scene understanding and a perspective branch for enhancing local detail and controllability.

Both branches leverage the UNet-based diffusion model from AnimateDiff~\cite{guo2023animatediff}, where each UNet block integrates a spatial Stable Diffusion (SD) UNet layer~\cite{rombach2022high} and a temporal motion module~\cite{guo2023animatediff} to learn spatial content and motion dynamics.

In the generation process, a single global text prompt guides the panorama branch to preserve overall scene coherence, while multiple local text prompts steer the perspective branch to enhance content diversity.
The two branches collaborate seamlessly to produce motion-rich and realistic panorama videos.

\textbf{Panorama Branch.}
The panorama branch is designed to provide global guidance, ensuring overall scene consistency and seamless boundary stitching in the generated panoramic view.
Give a global text description $y*$, the model generates a panoramic video $x*\in \mathbb{R}^{T \times 3 \times H \times W}$, where $W=2H$.
Due to the resolution mismatch between the pretrained SD backbone and the panoramic format, we finetune the spatial layers using LoRA\cite{hu2022lora} modules.
One challenge in panorama video generation is the curved motion patterns in equirectangular projections (ERPs) versus the straight-line motion in standard videos.
Thus, the motion module is finetuned to adapt the model to the curved pattern.
Moreover, to enforce seamless looping in panoramic views, we follow prior works \cite{fang2023ctrl, wu2023ipo} and apply circular padding\cite{shum2023conditional, wang2023360, zhuang2022acdnet} to the encoded video latent, as well as all feature maps before convolutional layers.
At inference time, we also apply a 90-degree latent rotation at each diffusion step to further mitigate boundary discontinuities.

\textbf{Perspective Branch.}
The objective of the perspective branch is to leverage the strong generative capabilities of the foundation video diffusion models to enhance diversity and realism in generated scenes.
We design this branch to simultaneously generate $N=4$ perspective videos $\{x^i\in \mathbb{R}^{T \times 3 \times H/2 \times H/2}\}_{i=1}^N$.
To ensure uniform spatial coverage, these views are uniformly distributed along the horizontal plane with azimuth angles of 0-, 90-, 180-, and 270-degree, all at 0-degree elevation and a 90-degree field of view (FOV).
As with the panorama branch, LoRA modules are employed to finetune the spatial layers for adaptation to the new resolution.

Previous research\cite{mao2023guided} suggests that noise initialization significantly influences the joint generation quality.
When multiple correlated views are generated concurrently, independently sampling noise can lead to scene inconsistencies. 
To maintain scene consistency, we adopt a shared noise initialization strategy~\cite{zhang2024taming}. 
Specifically, we derive the noise map for the perspective views by ERP-based projections from the noise map of the panorama view.
During training, noisy panorama video latents $z^{*}_{t} \in \mathbb{R}^{T \times C \times H/f \times W/f}$ are constructed with randomly sampled noise $\epsilon^{*}$ of the same dimension, where $C$ is the channel dimension and $f$ is the encoder spatial compression rate.
The noisy perspective video latents $\{z^i_t\in \mathbb{R}^{T \times C \times H/2f \times H/2f}\}_{i=1}^N$ are constructed by adding noise $\{\epsilon^{i}\}_{i=1}^N$ projected from $\epsilon^{*}$.
This same sampling strategy also applies during the inference stage.

\begin{figure}[t]
    \centering
    \includegraphics[width=0.98\linewidth]{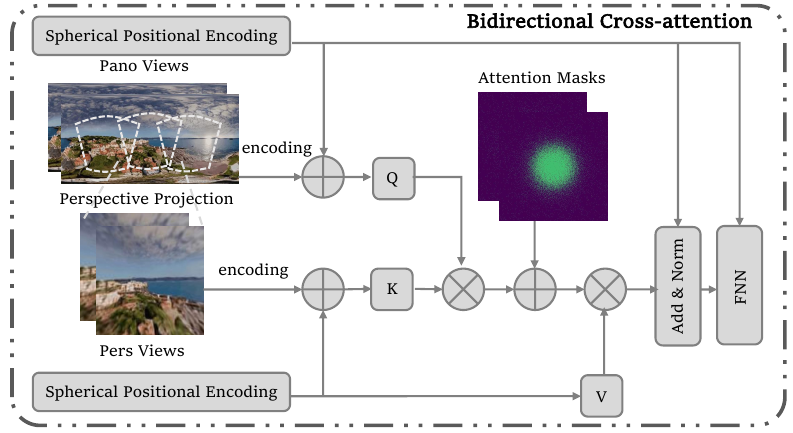}
    \caption{Bidirectional Cross-attention. We employ the bidirectional spatial-temporal cross-attention module to facilitate information transfer between the two branches. The diagram illustrates the single-direction information flow from the perspective branch to the panorama branch.}
    \label{fig:cross-attention}
\end{figure}

\textbf{Bidirectional Cross-attention.}
To effectively combine the global contextual guidance of the panorama branch with the fine-grained generation capability of the perspective branch, we introduce a bidirectional cross-attention module at each UNet layer.
As shown in Fig.~\ref{fig:cross-attention}, for each direction in the module, visual features from one branch act as queries while features from the other serve as keys and values; the bidirectional mechanism enables mutual information exchange.

Given the geometric differences between panorama and perspective formats, we incorporate spatial correspondence into the attention mechanism by applying Spherical Positional Encoding~\cite{zhang2024taming}.
This encoding projects positional information from the panorama onto the perspective feature map, enabling the model to align and integrate features across formats effectively.
Additionally, to ensure that attentions are applied only within the spatial-temporal corresponding regions between panoramic and perspective views, we introduce an ERP-guided attention mask over visual features from two branches, which emphasizes semantically relevant feature interactions and suppresses irrelevant regions.
Empirically, we find that using joint spatial-temporal cross-attention yields superior results than employing disjoint spatial and temporal cross-attention mechanisms.

\textbf{Progressive Training.}
We adopt a two-stage progressive learning strategy.
We start by training a dual-branch panorama image diffusion model.
Both branches are initialized from a pretrained SD UNet~\cite{rombach2022high}, where LoRA modules are inserted into spatial layers to adapt models to novel resolutions and broader content distribution.
During this phase, the bidirectional cross-attention modules operate only in the spatial dimension.
Subsequently, we extend model training for panoramic video generation by incorporating pretrained motion modules from~\cite{chen2024follow}.
In this phase, the spatial LoRA layers are frozen, while the bidirectional spatial-temporal cross-attention modules and motion modules are both finetuned to model curved motion trajectories in ERP space.
During both training stages, a global text prompt conditions the panorama branch, while four localized prompts guide the generation of corresponding perspective views.

For video generation, the training objectives are formulated as follows:
\begin{equation}
\mathcal{L}_{gen}=\mathcal{L}^* + \frac{1}{N} {\textstyle \sum_{i=1}^{N}}\mathcal{L}^i
\end{equation},
where
\begin{equation}
 \mathcal {L}^{*} = \mathbb {E}_{{\mathcal {E}}({x}^{*}),{t},{\epsilon }^{*},{y}} \left [ || {\epsilon }^{*} - {\epsilon }_{\theta }^{*}({z}_{t}^{*},{t},y^{*}) ||_2 \right ],
\end{equation}
\begin{equation}
 \mathcal {L}^{i} = \mathbb {E}_{{\mathcal {E}}({x}^{i}),{t},{\epsilon }^{i},{y}} \left [ || {\epsilon }^{i} - {\epsilon }_{\theta }^{i}({z}_{t}^{i},{t},y^{i}) ||_2 \right ]. 
\end{equation}
Here, ${\mathcal {E}}(x)$ denotes the encoding of the input data $x$, $y^{*}$ is the text description for the global panoramic view, and $y^{i}$ is the text description for $i$-th perspective view.

\subsection{Geometry-aligned Reconstruction Model}

Given the generated panoramic video, at this stage, we explicitly model the 4D scenes with Gaussian splatting to enable novel view synthesis and immersive viewing experience.

The source video consists of $T$ frames, denoted as $\{f^t\}_{t=1}^T$.

A straightforward approach would be to independently reconstruct a 3D Gaussian representation for each frame, resulting in $T$ separate sets of 3D Gaussians.
However, this approach overlooks the temporal continuity in the source video sequence and treats each frame in isolation, leading to geometry and motion inconsistency in the reconstructed scenes.
In particular, without constraints enforcing metric consistency, the geometry of each frame may drift across frames and compromise the scene coherence.
Furthermore, within a single panoramic frame, multiple perspective views should adhere to a shared geometric metric.

Here, we introduce a spatial-temporal alignment strategy to maintain consistency throughout the entire scene.
Spatial alignment enforces a shared geometric metric across all views within each frame, promoting geometric consistency across perspectives.
Temporal alignment further enhances coherence by aligning the reconstructed geometry over time, ensuring smooth and consistent scene evolution.
Additionally, we leverage camera pose estimation to initialize scene poses and alleviate inconsistencies introduced by camera motions.
We then perform per-scene optimization within this unified 4D field, effectively lifting the panorama video into an immersive and explicit scene representation.

\textbf{Spatial Alignment.}
We begin 4D scene reconstruction by initializing the scene geometry with dense spatial alignment.
Unlike prior methods that rely on sparse point clouds to construct 3D Gaussian Splatting (3DGS), we propose a dense initialization strategy by leveraging pixel-wise depth maps for each panoramic frame.

However, existing monocular depth estimation models are tailored for perspective views and degrade in handling panoramic images.
To address this, we project each panoramic frame into $K=20$ overlapping tangent perspective views of resolution $h_{sa} \times w_{sa}$ and estimate monocular depth maps $\{D^k\}_{k=1}^K$ using DPT\cite{ranftl2021vision}.
To obtain a unified panorama depth map, inspired by~\cite{rey2022360monodepth}, we spatially align the estimated perspective depths into a single coherent depth map. This requires estimating the scale and shift parameters for each perspective depth map ${D^k}$ to reconstruct a metric-aligned panorama depth.
Following~\cite{zhou2024dreamscene360}, we utilize a learnable global geometric field (MLPs) to regress the per-view scale and per-pixel shift parameters: $\left \{ \alpha _{k}\in R, \beta _{k} \in R^{h_{sa}\times w_{sa}} \right \} ^{K}_{k=1}$.

Here, ignoring $k$ without loss of generality, we define the optimization objective as:
\begin{equation}
    \min _{\alpha, \beta, \theta}\left\{ \mathcal{L}_{\mathrm{depth}}+\lambda_{\mathrm{\alpha }} \mathcal{L}_{\mathrm{\alpha }}+ \lambda_{\mathrm{\beta }} \mathcal{L}_{\mathrm{\beta }}   \right\}
\end{equation}
, where
\begin{equation}
    \mathcal{L}_{\mathrm{depth}}=|| \alpha \cdot D +  {\beta } - \operatorname {MLPs}( {v}; \theta ) ||_2^2,
\end{equation}
\begin{equation}
    \mathcal{L}_{\mathrm{\alpha}}=||\gamma (\alpha ) - 1||^2,
\end{equation}
\begin{equation}
    \mathcal{L}_{\mathrm{\beta}}=\sum _{i,j} \left ( (\beta _{i,j+1} - \beta _{i,j})^2 + (\beta _{i+1,j} - \beta _{i,j})^2 \right ).
\end{equation}

Here, $\operatorname {MLPs}$ parameterized by $\theta$ learn a global geometric field with view direction $v$ as input.

 $\mathcal{L}_{\mathrm{\alpha}}$ and  $\mathcal{L}_{\mathrm{\beta}}$ are regularization terms for $\alpha$ and $\beta$ with coefficients $\lambda_{\mathrm{\alpha }}$ and $\lambda_{\mathrm{\beta }}$.
 $\gamma \left ( \cdot \right )$ is the softplus function.

 \begin{figure*}[ht]
    \centering
    \includegraphics[width=\textwidth]{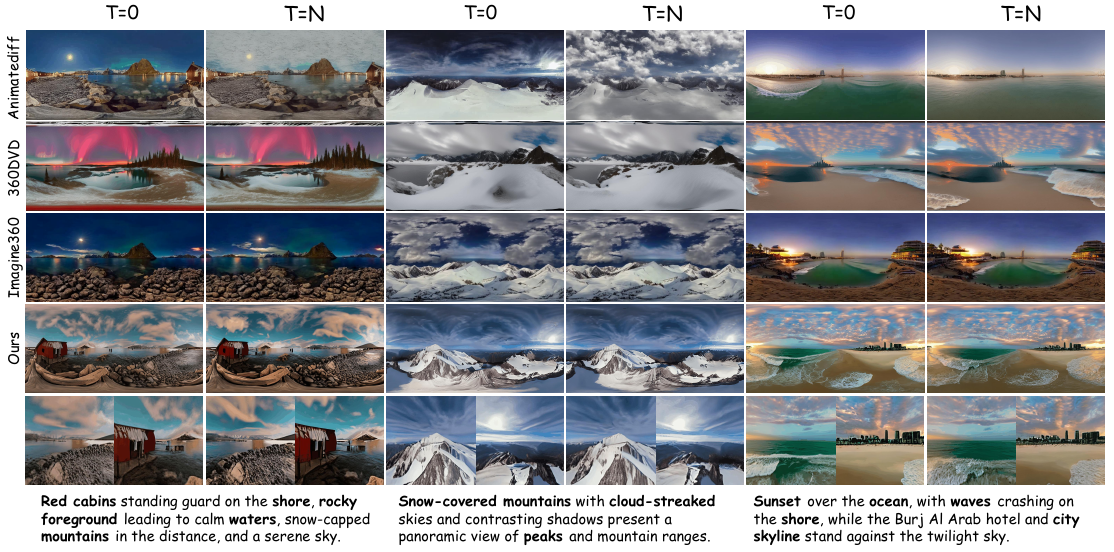}
    \caption{Qualitative Comparison of Panorama Video Generation. For our method, we show panoramic views and two perspective views. Our method generates motion-rich, semantically diverse scenes with fine-grained control and superior visual quality.
    }
    \label{fig:Qualitative Comparison of Video Generation}
\end{figure*}

\textbf{Temporal Alignment.}
After spatial alignment, we initialize the geometry for each frame using a unified panorama depth map.
However, along the temporal dimension, the geometry of individual frames remains independent and the panoramic depth maps exist at different scales, which can cause an inconsistency issue during novel view rendering. 
To align all depth maps within a consistent metric space, we project a single perspective view from the center of the panorama video and apply Monst3r\cite{zhang2024monst3r} to obtain a unified sequence of metric depth maps $\{{d}^{t}_{metric}\}_{t=1}^T$ along with the corresponding camera poses $\{{P}^{t}\}_{t=1}^T$.
Using these metric depth maps as supervision, we calibrate the panoramic depth sequence $\{ \boldsymbol{D}^t\}_{t=1}^T$ with per-view scale and shift parameters: $\left\{ \boldsymbol{\alpha}^t , \boldsymbol{\beta}^t \in \mathbb{R} \right\}_{t=1}^{T}$.

The calibration procedure is formulated as follows:
\begin{equation}
    \boldsymbol{D}^t_{aligned} = \boldsymbol{\alpha}^t \cdot \boldsymbol{D}^t + \boldsymbol{\beta}^t,
\end{equation}

 where,
 \begin{equation}
  \boldsymbol{\alpha}^t = \varphi(\dfrac{  d^{t}_{metric}}{{d}^t}),
 \end{equation}
\begin{equation}
  \boldsymbol{\beta}^t = \varphi(d^{t}_{metric}-\boldsymbol{\alpha}^t \cdot {d}^{t}).
 \end{equation}
Here, ${d}^{t}$ presents the perspective depth map projected from unaligned panoramic depth $\boldsymbol{D}^{t}$, $\varphi(\cdot)$ denotes the median operation.

\textbf{Training with estimated cameras.}
Following the geometry initialization and spatial-temporal alignment of the point clouds, we obtain an initial consistent scene geometry.
 Initializing all scene cameras at the same position would overlook the camera movements in the video, resulting in inconsistency in the reconstructed scene.

Therefore, we use the camera poses $\{{P}^{t}\}_{t=1}^T$ estimated from the previous step to initialize the scene cameras. 
We then apply 3D Gaussian Splatting to reconstruct the scene. 
The positions of the 3D Gaussians are initialized from the aligned point clouds, and the entire 4D scene is represented as $\operatorname{T}$ sets of 3D Gaussians.

Following~\cite{zhou2024dreamscene360}, these Gaussians are optimized with the objective:
\begin{equation}
\mathcal{L}_{recon}=\mathcal{L}_{\mathrm{rgb}}+\lambda_{\mathrm{sem}} \mathcal{L}_{\mathrm{sem}}+ \lambda_{\mathrm{geo}} \mathcal{L}_{\mathrm{geo}}.
\end{equation}
Here, the RGB loss is defined as:
\begin{equation}
\mathcal{L}_{\mathrm{rgb}}= \lambda_1 \mathcal{L}_{1}+\lambda_{ssim}\mathcal{L}_{\operatorname{ssim}}+\lambda_{\mathrm{lpips}}\mathcal{L}_{\mathrm{lpips}}.
\end{equation}
The distilling semantic loss is given by:
\begin{equation}
\mathcal {L}_{\textit {sem}} = 1 - \operatorname {cos}(\text {cls}(I_t), \text {cls}(I'_t))
\end{equation}
, where $\operatorname{cos}(\cdot, \cdot)$ represents cosine similarity and $\operatorname{cls}(\cdot)$ denotes feature extraction using DINOv2~\cite{oquab2023dinov2}. Here, $I_t$ refers to the rendered perspective image using camera viewport $P^t$, while $I'_t$ is obtained by slightly perturbing the camera $P^t$.

The geometric correspondence loss is defined as:
\begin{equation}
\mathcal {L}_\textit {geo} = 1 - \frac {\text {Cov}(D^t, \text {DPT}(I_t))}{\sqrt {\text {Var}(D^t)\text {Var}(\text {DPT}(I_t))}}
\end{equation}
where $D^t$ is the rendered depth map and $\operatorname{DPT}(\cdot)$ denotes monocular depth estimation~\cite{ranftl2021vision} of the input image. $\lambda$ terms are loss weights.

\begin{figure*}[ht]
    \centering
    \includegraphics[width=\textwidth]{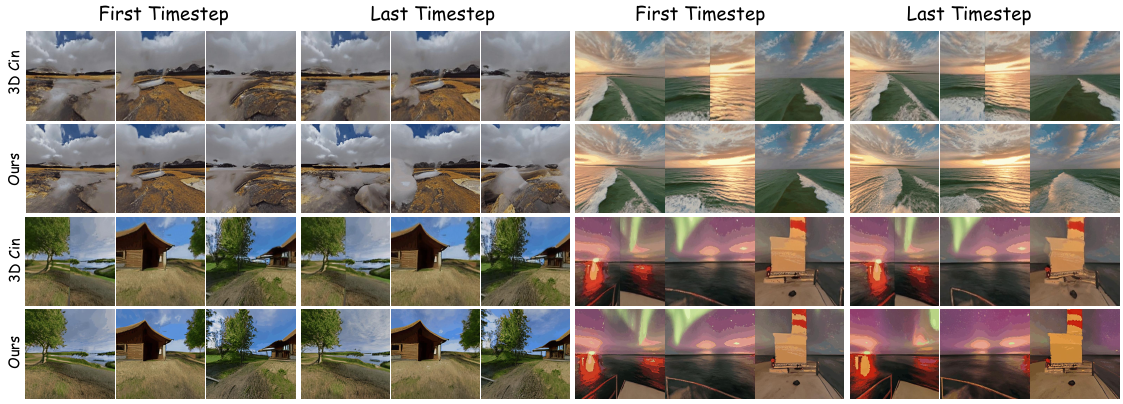}
    \caption{Qualitative Comparison of Reconstructed Scenes. We show rendered images from multiple perspective views at the first and last timestamps. Our model surpasses prior works in reconstructing geometry-consistent and visually realistic scenes.}
    \label{fig:Qualitative Comparison of Reconstructed Scenes}
\end{figure*}

\begin{table*}[!h]
\small
  \caption{Quantitative comparisons with state-of-the-art methods on video generation and 4D reconstruction. Our method consistently outperforms prior works across all metrics for text-visual alignment and visual quality.}
  \label{tab:Quantitative Comparisons}
  \begin{tabular}{lcccccccc}
    \toprule
    \textbf{Method} & \textbf{FID} $\downarrow$ & \textbf{FVD} $\downarrow$ & \textbf{CLIP} $\uparrow$ & \textbf{Q-Align (VQ)} $\uparrow$ & \makecell{\textbf{Imaging}\\\textbf{Quality} $\uparrow$} & \makecell{\textbf{Aesthetic}\\\textbf{Quality} $\uparrow$} & \makecell{\textbf{Overall}\\\textbf{Consistency} $\uparrow$} & \makecell{\textbf{Motion}\\\textbf{Smoothness} $\uparrow$} \\
    \midrule
    \multicolumn{9}{l}{\textbf{Video Generation Models}} \\
    AnimateDiff~\cite{guo2023animatediff} & 84.394 & 91.593 & 0.276 & 0.383 & 0.506 & 0.445 & 0.086 & 0.974 \\
    360DVD~\cite{wang2024360dvd} & 79.991 & 95.772 & 0.285 & 0.487 & 0.456 & 0.489 & 0.093 & 0.969 \\
    Imagine360~\cite{tan2024imagine360} & 82.429 & 89.851 & 0.279 & 0.492 & 0.591 & 0.473 & 0.095 & 0.979 \\
    \textbf{Ours-Video} & \textbf{70.802} & \textbf{74.963} & \textbf{0.296} & \textbf{0.501} & \textbf{0.613} & \textbf{0.557} & \textbf{0.134} & \textbf{0.984} \\
    \midrule
    \multicolumn{9}{l}{\textbf{Reconstruction Models}} \\
    3D-Cin~\cite{li20233d} & 79.446 & 93.990 & 0.278 & 0.340 & 0.440 & 0.466 & 0.088 & 0.972 \\
    \textbf{Ours-4D} & \textbf{75.641} & \textbf{76.542} & \textbf{0.289} & \textbf{0.517} & \textbf{0.620} & \textbf{0.496} & \textbf{0.102} & \textbf{0.996} \\
    \bottomrule
  \end{tabular}

\end{table*}

\section{Experiments}

\subsection{Experimental Setup}
\textbf{Dataset.} We use the Matterport3D~\cite{chang2017matterport3d} and WEB360~\cite{wang2024360dvd} datasets to train our dual-branch panorama generation model.
Matterport3D provides 10,800 high-quality panoramic images and is used for initial panoramic image diffusion model training.
WEB360 contains 2,114 high-resolution ERP-format text-video pairs and is used for both image and video model training.
Both datasets include global text descriptions. We employ CogVLM~\cite{wang2023cogvlm, hong2023cogagent} model to caption the perspective views and obtain the local text prompts.

\textbf{Baselines.}
We evaluate our method from two perspectives: panorama video generation and explicit 4D panorama scene reconstruction. 
We leave out 100 samples from WEB360 as the evaluation set.
For panorama video generation, we compare our model with state-of-the-art methods including AnimateDiff~\cite{guo2023animatediff}, 360DVD~\cite{wang2024360dvd}, and Imagine360~\cite{tan2024imagine360}.
360DVD is an advanced text-to-panorama video generation method and aligns well with our experimental setting.
For AnimateDiff, we adopt the LatentLab360 LoRA and pretrained motion modules to animate the conditional images derived from the ground truth panorama frames in the test set.
Imagine360 is a perspective-to-panoramic video generation framework, and we use a sampled perspective video from ground truth panorama videos as prompts.
For 4D panorama scene reconstruction, we compare with 3D-Cinemagraphy (3D-Cin)~\cite{li20233d}, an optical flow-based approach for dynamic 3D image creation.

\textbf{Evaluation Metrics.}
For panoramic video generation, we first assess the visual quality and textual alignment using Fréchet Inception Distance (FID)~\cite{heusel2017gans}, Fréchet Video Distance (FVD)~\cite{unterthiner2018towards}, and CLIP~\cite{hessel2021clipscore} score.
Following~\cite{tan2024imagine360}, we evaluate overall video quality using Video Quality Scorer (VQ) from Q-Align~\cite{wu2023q}. We also project panoramic videos into four perspective videos and evaluate graphics quality and motion consistency using metrics from VBench~\cite{huang2023vbench}, including Imaging Quality, Aesthetic Quality, Overall Consistency, and Motion Smoothness.
For the reconstruction model, we evaluate perspective videos rendered from novel viewpoints using the same set of metrics. For FID and FVD computation, ground-truth panoramic videos are projected into corresponding perspective views and used as reference targets.

\subsection{Comparisons of Video Generation}
For panorama video generation, we compare our method against Animatediff\cite{guo2023animatediff}, 360DVD\cite{wang2024360dvd}, and Imagine360\cite{tan2024imagine360}.
As shown in Fig. \ref{fig:Qualitative Comparison of Video Generation}, our dual-branch panorama video generation model exhibits more coherent and richer global content. In contrast, the outputs of AnimateDiff and 360DVD tend to be overly simplistic and exhibit weak alignment with the corresponding text prompts.
The videos generated by Imagine360 display minimal motion, while noticeable artifacts appear in the top and bottom regions of outputs from AnimateDiff.
In perspective views, our model provides finer details, superior visual fidelity, and smoother, more natural camera motion. It also achieves better alignment with the provided text prompts, showcasing its ability to support fine-grained control over local content.
For the quantitative evaluation, as shown in the upper section of Table~\ref{tab:Quantitative Comparisons}, our model consistently outperforms the baselines across all evaluation metrics, confirming its advantages in visual quality, content richness, motion dynamics, and temporal coherence.\label{para:4.2}

 \begin{figure}[t]
    \centering
    \includegraphics[width=\linewidth]{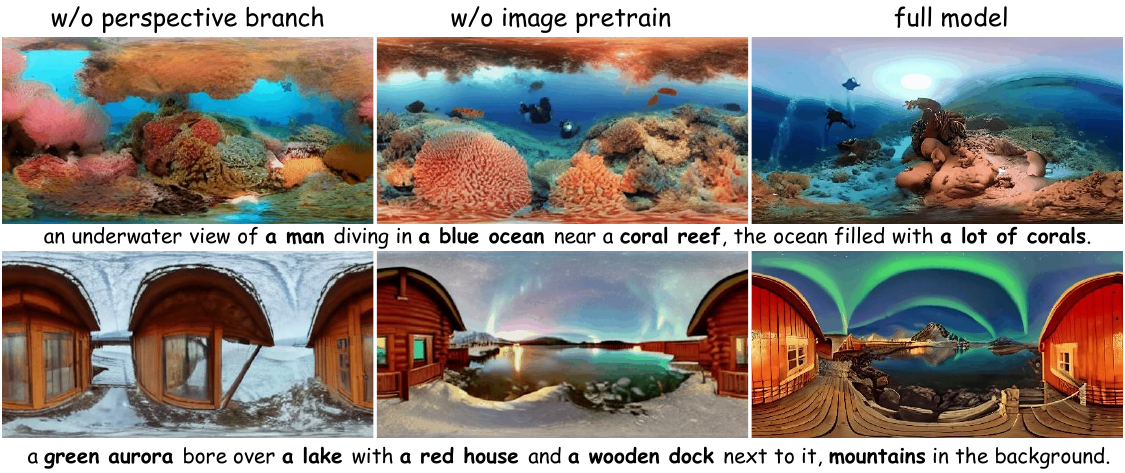}
    \caption{Visualizations of Video Generation Ablations. Compared with models without perspective branch or image diffusion pretraining, our full model generates videos with fine-grained control over contents and superior visual quality.}
    \label{fig:Ablation Studies of Video Generation}
\end{figure}

\begin{table}[t]
\footnotesize
    \centering
  \caption{Quantitative Results of Ablation Studies over Dual-branch Video Generation and Reconstruction Model.}
  \label{tab:Ablation Studies}
  \begin{tabular}{ccccc}
    \toprule
     Method & FID  $\downarrow$ &  VQ $\uparrow$ & \makecell{Imaging\\ Quality} $\uparrow$ & \makecell{Aesthetic\\ Quality} $\uparrow$ \\
    \midrule
     \textbf{Video Generation Model} \\
    w/o perspective branch & 102.722 & 0.458 & 0.541 & 0.452 \\
     w/o image pretrain & 95.416 & 0.462 & 0.522 & 0.475 \\
     full model & \textbf{70.802} & \textbf{0.501} & \textbf{0.613} & \textbf{0.557}\\
     \midrule
     \textbf{Reconstruction Model} \\
     w/o spatial alignment  & 136.080 & 0.447 & 0.529 & 0.427\\
     w/o temporal alignment & 129.519 & 0.463 & 0.578 & 0.452\\
      w/o camera poses & 132.116 & 0.457 & 0.575 & 0.433\\
     full model & \textbf{75.641} & \textbf{0.517} & \textbf{0.620} & \textbf{0.496}\\
  \bottomrule
\end{tabular}
\end{table}

\subsection{Comparisons of Reconstructed Scenes}
We also conduct both qualitative and quantitative analyses for reconstructed scene evaluation. To simulate novel viewpoints, we introduce perturbations to the training camera poses to generate test camera trajectories. These test poses are then used to render videos from the reconstructed scenes. As shown in Fig.~\ref{fig:Qualitative Comparison of Reconstructed Scenes}, our method delivers superior visual fidelity and more realistic scene motion compared to 3D-Cin. The 3D-Cin results exhibit minimal motion and visible discontinuities at scene transitions. Additionally, objects appear more blurred and distorted, compromising overall scene quality.
In contrast, our method produces sharper details, smoother motion, and more coherent geometry. As reported in Table~\ref{tab:Quantitative Comparisons}, our model consistently outperforms 3D-Cin across all evaluation metrics, highlighting its stronger generalization ability and improved reconstruction fidelity from novel viewpoints.

\subsection{Ablation Studies}
For dual-branch video generation model, we conduct ablation studies a) on model architecture by using panoramic branch only, and b) on training strategy by skipping the image diffusion model training phrase and directly training a dual-branch panorama video model, where the spatial LoRA modules and temporal motion modules are optimized altogether.
Results in Fig.~\ref{fig:Ablation Studies of Video Generation} and Tab.~\ref{tab:Ablation Studies} show that removing the perspective branch results in a loss of fine-grained control over the generated contents.
Skipping image model training leads to a noticeable degradation in visual quality.

\begin{figure}[t]
    \centering
    \includegraphics[width=\linewidth]{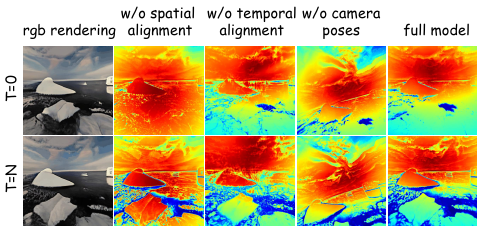 }
    \caption{Ablation Studies for Scene Reconstruction. Using full model obtains more consistent renderings and more reasonable geometric structures compared to models without spatial alignment or temporal alignment.}
    \label{fig:Ablation Studies of Reconstructed Scenes}
\end{figure}

We performed ablation studies on spatial and temporal alignments and camera pose estimation for 4D reconstruction. As shown in Fig. \ref{fig:Ablation Studies of Reconstructed Scenes}, our full model produces smoother, more coherent depth maps, indicating superior geometric structure. Removing spatial alignment or pose estimation leads to disordered depth, while omitting temporal alignment results in discontinuous geometry. These qualitative findings are supported by quantitative results in Tab. \ref{tab:Ablation Studies}, where the full model consistently outperforms all variants.

\begin{table}[t]
\scriptsize
\centering
\caption{User study results for dual-branch video generation and 4D reconstruction. We evaluate the criteria: Geometry Consistency (GC), Appearance Quality (AQ), Text Alignment (TA), Motion Magnitude (MM), and Content Richness (CR) for general video quality; and End Continuity (EC) and Motion Pattern (MP) for panorama-specific aspects. Our method consistently achieves the highest scores across all criteria.}
\begin{tabular}{cccccccc}
\toprule
 & \multicolumn{5}{c}{\textbf{Video Criteria}} & \multicolumn{2}{c}{\textbf{Panorama Criteria}} \\
\cmidrule(lr){2-6} \cmidrule(lr){7-8}
\textbf{Method} & \textbf{GC} $\uparrow$ & \textbf{AQ} $\uparrow$ & \textbf{TA} $\uparrow$ & \textbf{MM} $\uparrow$ & \textbf{CR} $\uparrow$ & \textbf{EC} $\uparrow$ & \textbf{MP} $\uparrow$ \\
\midrule
    Animatediff\cite{guo2023animatediff}& 3.3 & 3.4 & 3.3 & 3.2 &3.1 & 3.7 & 3.9\\
    360DVD\cite{wang2024360dvd}& 3.4 & 3.2 & 3.5 & 3.7 &3.5 & 3.5 & 3.6\\
    Imagine360\cite{tan2024imagine360} & 3.9 & 4.0 & 3.6 & 3.8 & 3.3 & 4.0 & 3.5 \\
    Ours-Video & \textbf{4.2} & \textbf{4.5} & \textbf{4.3} & \textbf{4.6} &\textbf{4.3} & \textbf{4.9} & \textbf{4.7}\\
    \midrule
     3D-Cin\cite{li20233d} & 3.1 & 3.3 & 3.0 & 3.0 &2.9 & -- & -- \\
    Ours-4D & \textbf{4.0} & \textbf{4.2} & \textbf{3.9} & \textbf{4.5} & \textbf{4.0} & --- & -- \\
\bottomrule
\end{tabular}
    \vspace{-0.2cm}
\label{tab:User Studies}
\end{table}

\subsection{User Studies}
We conducted user studies with 20 participants across 20 test scenes to subjectively evaluate our method. Participants rated general video quality (Geometry Consistency, Appearance Quality, Text Alignment, Motion Magnitude, Content Richness) and panorama-specific criteria (End Continuity, Motion Pattern) on a 1-to-5 scale. Our method consistently achieved the highest average scores across all metrics (Tab. \ref{tab:User Studies}), demonstrating significant improvements in image and geometry quality, textual alignment, scene content richness, panoramic consistency, and motion plausibility.

\section{Conclusion}
We introduce TiP4GEN, a novel framework for generating high-quality, immersive, dynamic panorama scenes directly from text. It features a dual-branch architecture for panorama video generation, ensuring both global coherence and local diversity. Subsequently, spatial-temporal alignment reconstructs these videos into geometrically consistent, interactive 4D scene representations. Experiments confirm TiP4GEN's superior ability to create visually compelling and motion-coherent dynamic panoramic scenes.

\begin{acks}
This research was funded by the Fundamental Research Funds for the Central Universities (2024XKRC082), the National Natural Science Foundation of China (No.92470203) and Beijing Natural Science Foundation (No. L242022).
\end{acks}

\bibliographystyle{ACM-Reference-Format}
\balance
\bibliography{sample-sigconf-authordraft}


\begin{thebibliography}{65}


\ifx \showCODEN    \undefined \def \showCODEN     #1{\unskip}     \fi
\ifx \showISBNx    \undefined \def \showISBNx     #1{\unskip}     \fi
\ifx \showISBNxiii \undefined \def \showISBNxiii  #1{\unskip}     \fi
\ifx \showISSN     \undefined \def \showISSN      #1{\unskip}     \fi
\ifx \showLCCN     \undefined \def \showLCCN      #1{\unskip}     \fi
\ifx \shownote     \undefined \def \shownote      #1{#1}          \fi
\ifx \showarticletitle \undefined \def \showarticletitle #1{#1}   \fi
\ifx \showURL      \undefined \def \showURL       {\relax}        \fi
\providecommand\bibfield[2]{#2}
\providecommand\bibinfo[2]{#2}
\providecommand\natexlab[1]{#1}
\providecommand\showeprint[2][]{arXiv:#2}

\bibitem[Akimoto et~al\mbox{.}(2022)]%
        {akimoto2022diverse}
\bibfield{author}{\bibinfo{person}{Naofumi Akimoto}, \bibinfo{person}{Yuhi Matsuo}, {and} \bibinfo{person}{Yoshimitsu Aoki}.} \bibinfo{year}{2022}\natexlab{}.
\newblock \showarticletitle{Diverse plausible 360-degree image outpainting for efficient 3dcg background creation}. In \bibinfo{booktitle}{\emph{Proceedings of the IEEE/CVF Conference on Computer Vision and Pattern Recognition}}. \bibinfo{pages}{11441--11450}.
\newblock


\bibitem[Blattmann et~al\mbox{.}(2023a)]%
        {blattmann2023stable}
\bibfield{author}{\bibinfo{person}{Andreas Blattmann}, \bibinfo{person}{Tim Dockhorn}, \bibinfo{person}{Sumith Kulal}, \bibinfo{person}{Daniel Mendelevitch}, \bibinfo{person}{Maciej Kilian}, \bibinfo{person}{Dominik Lorenz}, \bibinfo{person}{Yam Levi}, \bibinfo{person}{Zion English}, \bibinfo{person}{Vikram Voleti}, \bibinfo{person}{Adam Letts}, {et~al\mbox{.}}} \bibinfo{year}{2023}\natexlab{a}.
\newblock \showarticletitle{Stable video diffusion: Scaling latent video diffusion models to large datasets}.
\newblock \bibinfo{journal}{\emph{arXiv preprint arXiv:2311.15127}} (\bibinfo{year}{2023}).
\newblock


\bibitem[Blattmann et~al\mbox{.}(2023b)]%
        {blattmann2023align}
\bibfield{author}{\bibinfo{person}{Andreas Blattmann}, \bibinfo{person}{Robin Rombach}, \bibinfo{person}{Huan Ling}, \bibinfo{person}{Tim Dockhorn}, \bibinfo{person}{Seung~Wook Kim}, \bibinfo{person}{Sanja Fidler}, {and} \bibinfo{person}{Karsten Kreis}.} \bibinfo{year}{2023}\natexlab{b}.
\newblock \showarticletitle{Align your latents: High-resolution video synthesis with latent diffusion models}. In \bibinfo{booktitle}{\emph{Proceedings of the IEEE/CVF conference on computer vision and pattern recognition}}. \bibinfo{pages}{22563--22575}.
\newblock


\bibitem[Chang et~al\mbox{.}(2017)]%
        {chang2017matterport3d}
\bibfield{author}{\bibinfo{person}{Angel Chang}, \bibinfo{person}{Angela Dai}, \bibinfo{person}{Thomas Funkhouser}, \bibinfo{person}{Maciej Halber}, \bibinfo{person}{Matthias Niessner}, \bibinfo{person}{Manolis Savva}, \bibinfo{person}{Shuran Song}, \bibinfo{person}{Andy Zeng}, {and} \bibinfo{person}{Yinda Zhang}.} \bibinfo{year}{2017}\natexlab{}.
\newblock \showarticletitle{Matterport3d: Learning from rgb-d data in indoor environments}.
\newblock \bibinfo{journal}{\emph{arXiv preprint arXiv:1709.06158}} (\bibinfo{year}{2017}).
\newblock


\bibitem[Chen et~al\mbox{.}(2024)]%
        {chen2024follow}
\bibfield{author}{\bibinfo{person}{Qihua Chen}, \bibinfo{person}{Yue Ma}, \bibinfo{person}{Hongfa Wang}, \bibinfo{person}{Junkun Yuan}, \bibinfo{person}{Wenzhe Zhao}, \bibinfo{person}{Qi Tian}, \bibinfo{person}{Hongmei Wang}, \bibinfo{person}{Shaobo Min}, \bibinfo{person}{Qifeng Chen}, {and} \bibinfo{person}{Wei Liu}.} \bibinfo{year}{2024}\natexlab{}.
\newblock \showarticletitle{Follow-your-canvas: Higher-resolution video outpainting with extensive content generation}.
\newblock \bibinfo{journal}{\emph{arXiv preprint arXiv:2409.01055}} (\bibinfo{year}{2024}).
\newblock


\bibitem[Chung et~al\mbox{.}(2023)]%
        {chung2023luciddreamer}
\bibfield{author}{\bibinfo{person}{Jaeyoung Chung}, \bibinfo{person}{Suyoung Lee}, \bibinfo{person}{Hyeongjin Nam}, \bibinfo{person}{Jaerin Lee}, {and} \bibinfo{person}{Kyoung~Mu Lee}.} \bibinfo{year}{2023}\natexlab{}.
\newblock \showarticletitle{Luciddreamer: Domain-free generation of 3d gaussian splatting scenes}.
\newblock \bibinfo{journal}{\emph{arXiv preprint arXiv:2311.13384}} (\bibinfo{year}{2023}).
\newblock


\bibitem[Dastjerdi et~al\mbox{.}(2022)]%
        {dastjerdi2022guided}
\bibfield{author}{\bibinfo{person}{Mohammad Reza~Karimi Dastjerdi}, \bibinfo{person}{Yannick Hold-Geoffroy}, \bibinfo{person}{Jonathan Eisenmann}, \bibinfo{person}{Siavash Khodadadeh}, {and} \bibinfo{person}{Jean-Fran{\c{c}}ois Lalonde}.} \bibinfo{year}{2022}\natexlab{}.
\newblock \showarticletitle{Guided co-modulated gan for 360 field of view extrapolation}. In \bibinfo{booktitle}{\emph{2022 International Conference on 3D Vision (3DV)}}. IEEE, \bibinfo{pages}{475--485}.
\newblock


\bibitem[Deng et~al\mbox{.}(2023)]%
        {deng2023mv}
\bibfield{author}{\bibinfo{person}{Zijun Deng}, \bibinfo{person}{Xiangteng He}, \bibinfo{person}{Yuxin Peng}, \bibinfo{person}{Xiongwei Zhu}, {and} \bibinfo{person}{Lele Cheng}.} \bibinfo{year}{2023}\natexlab{}.
\newblock \showarticletitle{MV-Diffusion: Motion-aware video diffusion model}. In \bibinfo{booktitle}{\emph{Proceedings of the 31st ACM International Conference on Multimedia}}. \bibinfo{pages}{7255--7263}.
\newblock


\bibitem[Fang et~al\mbox{.}(2023)]%
        {fang2023ctrl}
\bibfield{author}{\bibinfo{person}{Chuan Fang}, \bibinfo{person}{Yuan Dong}, \bibinfo{person}{Kunming Luo}, \bibinfo{person}{Xiaotao Hu}, \bibinfo{person}{Rakesh Shrestha}, {and} \bibinfo{person}{Ping Tan}.} \bibinfo{year}{2023}\natexlab{}.
\newblock \showarticletitle{Ctrl-Room: controllable text-to-3D room meshes generation with layout constraints}.
\newblock \bibinfo{journal}{\emph{arXiv preprint arXiv:2310.03602}} (\bibinfo{year}{2023}).
\newblock


\bibitem[Guo et~al\mbox{.}(2023)]%
        {guo2023animatediff}
\bibfield{author}{\bibinfo{person}{Yuwei Guo}, \bibinfo{person}{Ceyuan Yang}, \bibinfo{person}{Anyi Rao}, \bibinfo{person}{Zhengyang Liang}, \bibinfo{person}{Yaohui Wang}, \bibinfo{person}{Yu Qiao}, \bibinfo{person}{Maneesh Agrawala}, \bibinfo{person}{Dahua Lin}, {and} \bibinfo{person}{Bo Dai}.} \bibinfo{year}{2023}\natexlab{}.
\newblock \showarticletitle{Animatediff: Animate your personalized text-to-image diffusion models without specific tuning}.
\newblock \bibinfo{journal}{\emph{arXiv preprint arXiv:2307.04725}} (\bibinfo{year}{2023}).
\newblock


\bibitem[He et~al\mbox{.}(2022)]%
        {he2022latent}
\bibfield{author}{\bibinfo{person}{Yingqing He}, \bibinfo{person}{Tianyu Yang}, \bibinfo{person}{Yong Zhang}, \bibinfo{person}{Ying Shan}, {and} \bibinfo{person}{Qifeng Chen}.} \bibinfo{year}{2022}\natexlab{}.
\newblock \showarticletitle{Latent video diffusion models for high-fidelity long video generation}.
\newblock \bibinfo{journal}{\emph{arXiv preprint arXiv:2211.13221}} (\bibinfo{year}{2022}).
\newblock


\bibitem[Hessel et~al\mbox{.}(2021)]%
        {hessel2021clipscore}
\bibfield{author}{\bibinfo{person}{Jack Hessel}, \bibinfo{person}{Ari Holtzman}, \bibinfo{person}{Maxwell Forbes}, \bibinfo{person}{Ronan~Le Bras}, {and} \bibinfo{person}{Yejin Choi}.} \bibinfo{year}{2021}\natexlab{}.
\newblock \showarticletitle{Clipscore: A reference-free evaluation metric for image captioning}.
\newblock \bibinfo{journal}{\emph{arXiv preprint arXiv:2104.08718}} (\bibinfo{year}{2021}).
\newblock


\bibitem[Heusel et~al\mbox{.}(2017)]%
        {heusel2017gans}
\bibfield{author}{\bibinfo{person}{Martin Heusel}, \bibinfo{person}{Hubert Ramsauer}, \bibinfo{person}{Thomas Unterthiner}, \bibinfo{person}{Bernhard Nessler}, {and} \bibinfo{person}{Sepp Hochreiter}.} \bibinfo{year}{2017}\natexlab{}.
\newblock \showarticletitle{Gans trained by a two time-scale update rule converge to a local nash equilibrium}.
\newblock \bibinfo{journal}{\emph{Advances in neural information processing systems}}  \bibinfo{volume}{30} (\bibinfo{year}{2017}).
\newblock


\bibitem[Ho et~al\mbox{.}(2020)]%
        {ho2020denoising}
\bibfield{author}{\bibinfo{person}{Jonathan Ho}, \bibinfo{person}{Ajay Jain}, {and} \bibinfo{person}{Pieter Abbeel}.} \bibinfo{year}{2020}\natexlab{}.
\newblock \showarticletitle{Denoising diffusion probabilistic models}.
\newblock \bibinfo{journal}{\emph{Advances in neural information processing systems}}  \bibinfo{volume}{33} (\bibinfo{year}{2020}), \bibinfo{pages}{6840--6851}.
\newblock


\bibitem[Hong et~al\mbox{.}(2022)]%
        {hong2022cogvideo}
\bibfield{author}{\bibinfo{person}{Wenyi Hong}, \bibinfo{person}{Ming Ding}, \bibinfo{person}{Wendi Zheng}, \bibinfo{person}{Xinghan Liu}, {and} \bibinfo{person}{Jie Tang}.} \bibinfo{year}{2022}\natexlab{}.
\newblock \showarticletitle{Cogvideo: Large-scale pretraining for text-to-video generation via transformers}.
\newblock \bibinfo{journal}{\emph{arXiv preprint arXiv:2205.15868}} (\bibinfo{year}{2022}).
\newblock


\bibitem[Hong et~al\mbox{.}(2023)]%
        {hong2023cogagent}
\bibfield{author}{\bibinfo{person}{Wenyi Hong}, \bibinfo{person}{Weihan Wang}, \bibinfo{person}{Qingsong Lv}, \bibinfo{person}{Jiazheng Xu}, \bibinfo{person}{Wenmeng Yu}, \bibinfo{person}{Junhui Ji}, \bibinfo{person}{Yan Wang}, \bibinfo{person}{Zihan Wang}, \bibinfo{person}{Yuxiao Dong}, \bibinfo{person}{Ming Ding}, {and} \bibinfo{person}{Jie Tang}.} \bibinfo{year}{2023}\natexlab{}.
\newblock \bibinfo{title}{CogAgent: A Visual Language Model for GUI Agents}.
\newblock
\showeprint[arxiv]{2312.08914}~[cs.CV]


\bibitem[Hu et~al\mbox{.}(2022)]%
        {hu2022lora}
\bibfield{author}{\bibinfo{person}{Edward~J Hu}, \bibinfo{person}{Yelong Shen}, \bibinfo{person}{Phillip Wallis}, \bibinfo{person}{Zeyuan Allen-Zhu}, \bibinfo{person}{Yuanzhi Li}, \bibinfo{person}{Shean Wang}, \bibinfo{person}{Lu Wang}, \bibinfo{person}{Weizhu Chen}, {et~al\mbox{.}}} \bibinfo{year}{2022}\natexlab{}.
\newblock \showarticletitle{Lora: Low-rank adaptation of large language models.}
\newblock \bibinfo{journal}{\emph{ICLR}} \bibinfo{volume}{1}, \bibinfo{number}{2} (\bibinfo{year}{2022}), \bibinfo{pages}{3}.
\newblock


\bibitem[Huang et~al\mbox{.}(2024)]%
        {huang2023vbench}
\bibfield{author}{\bibinfo{person}{Ziqi Huang}, \bibinfo{person}{Yinan He}, \bibinfo{person}{Jiashuo Yu}, \bibinfo{person}{Fan Zhang}, \bibinfo{person}{Chenyang Si}, \bibinfo{person}{Yuming Jiang}, \bibinfo{person}{Yuanhan Zhang}, \bibinfo{person}{Tianxing Wu}, \bibinfo{person}{Qingyang Jin}, \bibinfo{person}{Nattapol Chanpaisit}, \bibinfo{person}{Yaohui Wang}, \bibinfo{person}{Xinyuan Chen}, \bibinfo{person}{Limin Wang}, \bibinfo{person}{Dahua Lin}, \bibinfo{person}{Yu Qiao}, {and} \bibinfo{person}{Ziwei Liu}.} \bibinfo{year}{2024}\natexlab{}.
\newblock \showarticletitle{{VBench}: Comprehensive Benchmark Suite for Video Generative Models}. In \bibinfo{booktitle}{\emph{Proceedings of the IEEE/CVF Conference on Computer Vision and Pattern Recognition}}.
\newblock


\bibitem[Karras et~al\mbox{.}(2022)]%
        {karras2022elucidating}
\bibfield{author}{\bibinfo{person}{Tero Karras}, \bibinfo{person}{Miika Aittala}, \bibinfo{person}{Timo Aila}, {and} \bibinfo{person}{Samuli Laine}.} \bibinfo{year}{2022}\natexlab{}.
\newblock \showarticletitle{Elucidating the design space of diffusion-based generative models}.
\newblock \bibinfo{journal}{\emph{Advances in neural information processing systems}}  \bibinfo{volume}{35} (\bibinfo{year}{2022}), \bibinfo{pages}{26565--26577}.
\newblock


\bibitem[Kerbl et~al\mbox{.}(2023)]%
        {kerbl3Dgaussians}
\bibfield{author}{\bibinfo{person}{Bernhard Kerbl}, \bibinfo{person}{Georgios Kopanas}, \bibinfo{person}{Thomas Leimk{\"u}hler}, {and} \bibinfo{person}{George Drettakis}.} \bibinfo{year}{2023}\natexlab{}.
\newblock \showarticletitle{3D Gaussian Splatting for Real-Time Radiance Field Rendering}.
\newblock \bibinfo{journal}{\emph{ACM Transactions on Graphics}} \bibinfo{volume}{42}, \bibinfo{number}{4} (\bibinfo{date}{July} \bibinfo{year}{2023}).
\newblock
\urldef\tempurl%
\url{https://repo-sam.inria.fr/fungraph/3d-gaussian-splatting/}
\showURL{%
\tempurl}


\bibitem[Lee et~al\mbox{.}(2024)]%
        {lee2024vividdream}
\bibfield{author}{\bibinfo{person}{Yao-Chih Lee}, \bibinfo{person}{Yi-Ting Chen}, \bibinfo{person}{Andrew Wang}, \bibinfo{person}{Ting-Hsuan Liao}, \bibinfo{person}{Brandon~Y Feng}, {and} \bibinfo{person}{Jia-Bin Huang}.} \bibinfo{year}{2024}\natexlab{}.
\newblock \showarticletitle{Vividdream: Generating 3d scene with ambient dynamics}.
\newblock \bibinfo{journal}{\emph{arXiv preprint arXiv:2405.20334}} (\bibinfo{year}{2024}).
\newblock


\bibitem[Li and Bansal(2023)]%
        {li2023panogen}
\bibfield{author}{\bibinfo{person}{Jialu Li} {and} \bibinfo{person}{Mohit Bansal}.} \bibinfo{year}{2023}\natexlab{}.
\newblock \showarticletitle{Panogen: Text-conditioned panoramic environment generation for vision-and-language navigation}.
\newblock \bibinfo{journal}{\emph{Advances in Neural Information Processing Systems}}  \bibinfo{volume}{36} (\bibinfo{year}{2023}), \bibinfo{pages}{21878--21894}.
\newblock


\bibitem[Li et~al\mbox{.}(2023b)]%
        {li2023blip}
\bibfield{author}{\bibinfo{person}{Junnan Li}, \bibinfo{person}{Dongxu Li}, \bibinfo{person}{Silvio Savarese}, {and} \bibinfo{person}{Steven Hoi}.} \bibinfo{year}{2023}\natexlab{b}.
\newblock \showarticletitle{Blip-2: Bootstrapping language-image pre-training with frozen image encoders and large language models}. In \bibinfo{booktitle}{\emph{International conference on machine learning}}. PMLR, \bibinfo{pages}{19730--19742}.
\newblock


\bibitem[Li et~al\mbox{.}(2024)]%
        {li20244k4dgen}
\bibfield{author}{\bibinfo{person}{Renjie Li}, \bibinfo{person}{Panwang Pan}, \bibinfo{person}{Bangbang Yang}, \bibinfo{person}{Dejia Xu}, \bibinfo{person}{Shijie Zhou}, \bibinfo{person}{Xuanyang Zhang}, \bibinfo{person}{Zeming Li}, \bibinfo{person}{Achuta Kadambi}, \bibinfo{person}{Zhangyang Wang}, \bibinfo{person}{Zhengzhong Tu}, {et~al\mbox{.}}} \bibinfo{year}{2024}\natexlab{}.
\newblock \showarticletitle{4k4dgen: Panoramic 4d generation at 4k resolution}.
\newblock \bibinfo{journal}{\emph{arXiv preprint arXiv:2406.13527}} (\bibinfo{year}{2024}).
\newblock


\bibitem[Li et~al\mbox{.}(2023a)]%
        {li20233d}
\bibfield{author}{\bibinfo{person}{Xingyi Li}, \bibinfo{person}{Zhiguo Cao}, \bibinfo{person}{Huiqiang Sun}, \bibinfo{person}{Jianming Zhang}, \bibinfo{person}{Ke Xian}, {and} \bibinfo{person}{Guosheng Lin}.} \bibinfo{year}{2023}\natexlab{a}.
\newblock \showarticletitle{3d cinemagraphy from a single image}. In \bibinfo{booktitle}{\emph{Proceedings of the IEEE/CVF Conference on Computer Vision and Pattern Recognition}}. \bibinfo{pages}{4595--4605}.
\newblock


\bibitem[Liang et~al\mbox{.}(2024a)]%
        {liang2024wonderland}
\bibfield{author}{\bibinfo{person}{Hanwen Liang}, \bibinfo{person}{Junli Cao}, \bibinfo{person}{Vidit Goel}, \bibinfo{person}{Guocheng Qian}, \bibinfo{person}{Sergei Korolev}, \bibinfo{person}{Demetri Terzopoulos}, \bibinfo{person}{Konstantinos~N Plataniotis}, \bibinfo{person}{Sergey Tulyakov}, {and} \bibinfo{person}{Jian Ren}.} \bibinfo{year}{2024}\natexlab{a}.
\newblock \showarticletitle{Wonderland: Navigating 3D Scenes from a Single Image}.
\newblock \bibinfo{journal}{\emph{arXiv preprint arXiv:2412.12091}} (\bibinfo{year}{2024}).
\newblock


\bibitem[Liang et~al\mbox{.}(2024b)]%
        {liang2024diffusion4d}
\bibfield{author}{\bibinfo{person}{Hanwen Liang}, \bibinfo{person}{Yuyang Yin}, \bibinfo{person}{Dejia Xu}, \bibinfo{person}{Hanxue Liang}, \bibinfo{person}{Zhangyang Wang}, \bibinfo{person}{Konstantinos~N Plataniotis}, \bibinfo{person}{Yao Zhao}, {and} \bibinfo{person}{Yunchao Wei}.} \bibinfo{year}{2024}\natexlab{b}.
\newblock \showarticletitle{Diffusion4d: Fast spatial-temporal consistent 4d generation via video diffusion models}.
\newblock \bibinfo{journal}{\emph{arXiv preprint arXiv:2405.16645}} (\bibinfo{year}{2024}).
\newblock


\bibitem[Liu et~al\mbox{.}(2023)]%
        {liu2023one}
\bibfield{author}{\bibinfo{person}{Minghua Liu}, \bibinfo{person}{Chao Xu}, \bibinfo{person}{Haian Jin}, \bibinfo{person}{Linghao Chen}, \bibinfo{person}{Mukund Varma~T}, \bibinfo{person}{Zexiang Xu}, {and} \bibinfo{person}{Hao Su}.} \bibinfo{year}{2023}\natexlab{}.
\newblock \showarticletitle{One-2-3-45: Any single image to 3d mesh in 45 seconds without per-shape optimization}.
\newblock \bibinfo{journal}{\emph{Advances in Neural Information Processing Systems}}  \bibinfo{volume}{36} (\bibinfo{year}{2023}), \bibinfo{pages}{22226--22246}.
\newblock


\bibitem[Liu et~al\mbox{.}(2024)]%
        {liu2024sora}
\bibfield{author}{\bibinfo{person}{Yixin Liu}, \bibinfo{person}{Kai Zhang}, \bibinfo{person}{Yuan Li}, \bibinfo{person}{Zhiling Yan}, \bibinfo{person}{Chujie Gao}, \bibinfo{person}{Ruoxi Chen}, \bibinfo{person}{Zhengqing Yuan}, \bibinfo{person}{Yue Huang}, \bibinfo{person}{Hanchi Sun}, \bibinfo{person}{Jianfeng Gao}, {et~al\mbox{.}}} \bibinfo{year}{2024}\natexlab{}.
\newblock \showarticletitle{Sora: A review on background, technology, limitations, and opportunities of large vision models}.
\newblock \bibinfo{journal}{\emph{arXiv preprint arXiv:2402.17177}} (\bibinfo{year}{2024}).
\newblock


\bibitem[Lu et~al\mbox{.}(2022)]%
        {NEURIPS2022_260a14ac}
\bibfield{author}{\bibinfo{person}{Cheng Lu}, \bibinfo{person}{Yuhao Zhou}, \bibinfo{person}{Fan Bao}, \bibinfo{person}{Jianfei Chen}, \bibinfo{person}{Chongxuan LI}, {and} \bibinfo{person}{Jun Zhu}.} \bibinfo{year}{2022}\natexlab{}.
\newblock \showarticletitle{DPM-Solver: A Fast ODE Solver for Diffusion Probabilistic Model Sampling in Around 10 Steps}. In \bibinfo{booktitle}{\emph{Advances in Neural Information Processing Systems}}, \bibfield{editor}{\bibinfo{person}{S.~Koyejo}, \bibinfo{person}{S.~Mohamed}, \bibinfo{person}{A.~Agarwal}, \bibinfo{person}{D.~Belgrave}, \bibinfo{person}{K.~Cho}, {and} \bibinfo{person}{A.~Oh}} (Eds.), Vol.~\bibinfo{volume}{35}. \bibinfo{publisher}{Curran Associates, Inc.}, \bibinfo{pages}{5775--5787}.
\newblock
\urldef\tempurl%
\url{https://proceedings.neurips.cc/paper_files/paper/2022/file/260a14acce2a89dad36adc8eefe7c59e-Paper-Conference.pdf}
\showURL{%
\tempurl}


\bibitem[Mao et~al\mbox{.}(2023)]%
        {mao2023guided}
\bibfield{author}{\bibinfo{person}{Jiafeng Mao}, \bibinfo{person}{Xueting Wang}, {and} \bibinfo{person}{Kiyoharu Aizawa}.} \bibinfo{year}{2023}\natexlab{}.
\newblock \showarticletitle{Guided image synthesis via initial image editing in diffusion model}. In \bibinfo{booktitle}{\emph{Proceedings of the 31st ACM International Conference on Multimedia}}. \bibinfo{pages}{5321--5329}.
\newblock


\bibitem[Nichol et~al\mbox{.}(2021)]%
        {nichol2021glide}
\bibfield{author}{\bibinfo{person}{Alex Nichol}, \bibinfo{person}{Prafulla Dhariwal}, \bibinfo{person}{Aditya Ramesh}, \bibinfo{person}{Pranav Shyam}, \bibinfo{person}{Pamela Mishkin}, \bibinfo{person}{Bob McGrew}, \bibinfo{person}{Ilya Sutskever}, {and} \bibinfo{person}{Mark Chen}.} \bibinfo{year}{2021}\natexlab{}.
\newblock \showarticletitle{Glide: Towards photorealistic image generation and editing with text-guided diffusion models}.
\newblock \bibinfo{journal}{\emph{arXiv preprint arXiv:2112.10741}} (\bibinfo{year}{2021}).
\newblock


\bibitem[Oquab et~al\mbox{.}(2023)]%
        {oquab2023dinov2}
\bibfield{author}{\bibinfo{person}{Maxime Oquab}, \bibinfo{person}{Timoth{\'e}e Darcet}, \bibinfo{person}{Th{\'e}o Moutakanni}, \bibinfo{person}{Huy Vo}, \bibinfo{person}{Marc Szafraniec}, \bibinfo{person}{Vasil Khalidov}, \bibinfo{person}{Pierre Fernandez}, \bibinfo{person}{Daniel Haziza}, \bibinfo{person}{Francisco Massa}, \bibinfo{person}{Alaaeldin El-Nouby}, {et~al\mbox{.}}} \bibinfo{year}{2023}\natexlab{}.
\newblock \showarticletitle{Dinov2: Learning robust visual features without supervision}.
\newblock \bibinfo{journal}{\emph{arXiv preprint arXiv:2304.07193}} (\bibinfo{year}{2023}).
\newblock


\bibitem[Podell et~al\mbox{.}(2023)]%
        {podell2023sdxlimprovinglatentdiffusion}
\bibfield{author}{\bibinfo{person}{Dustin Podell}, \bibinfo{person}{Zion English}, \bibinfo{person}{Kyle Lacey}, \bibinfo{person}{Andreas Blattmann}, \bibinfo{person}{Tim Dockhorn}, \bibinfo{person}{Jonas Müller}, \bibinfo{person}{Joe Penna}, {and} \bibinfo{person}{Robin Rombach}.} \bibinfo{year}{2023}\natexlab{}.
\newblock \bibinfo{title}{SDXL: Improving Latent Diffusion Models for High-Resolution Image Synthesis}.
\newblock
\showeprint[arxiv]{2307.01952}~[cs.CV]
\urldef\tempurl%
\url{https://arxiv.org/abs/2307.01952}
\showURL{%
\tempurl}


\bibitem[Poole et~al\mbox{.}(2022)]%
        {poole2022dreamfusion}
\bibfield{author}{\bibinfo{person}{Ben Poole}, \bibinfo{person}{Ajay Jain}, \bibinfo{person}{Jonathan~T Barron}, {and} \bibinfo{person}{Ben Mildenhall}.} \bibinfo{year}{2022}\natexlab{}.
\newblock \showarticletitle{Dreamfusion: Text-to-3d using 2d diffusion}.
\newblock \bibinfo{journal}{\emph{arXiv preprint arXiv:2209.14988}} (\bibinfo{year}{2022}).
\newblock


\bibitem[Ranftl et~al\mbox{.}(2021)]%
        {ranftl2021vision}
\bibfield{author}{\bibinfo{person}{Ren{\'e} Ranftl}, \bibinfo{person}{Alexey Bochkovskiy}, {and} \bibinfo{person}{Vladlen Koltun}.} \bibinfo{year}{2021}\natexlab{}.
\newblock \showarticletitle{Vision transformers for dense prediction}. In \bibinfo{booktitle}{\emph{Proceedings of the IEEE/CVF international conference on computer vision}}. \bibinfo{pages}{12179--12188}.
\newblock


\bibitem[Rey-Area et~al\mbox{.}(2022)]%
        {rey2022360monodepth}
\bibfield{author}{\bibinfo{person}{Manuel Rey-Area}, \bibinfo{person}{Mingze Yuan}, {and} \bibinfo{person}{Christian Richardt}.} \bibinfo{year}{2022}\natexlab{}.
\newblock \showarticletitle{360monodepth: High-resolution 360deg monocular depth estimation}. In \bibinfo{booktitle}{\emph{Proceedings of the IEEE/CVF Conference on Computer Vision and Pattern Recognition}}. \bibinfo{pages}{3762--3772}.
\newblock


\bibitem[Rombach et~al\mbox{.}(2022)]%
        {rombach2022high}
\bibfield{author}{\bibinfo{person}{Robin Rombach}, \bibinfo{person}{Andreas Blattmann}, \bibinfo{person}{Dominik Lorenz}, \bibinfo{person}{Patrick Esser}, {and} \bibinfo{person}{Bj{\"o}rn Ommer}.} \bibinfo{year}{2022}\natexlab{}.
\newblock \showarticletitle{High-resolution image synthesis with latent diffusion models}. In \bibinfo{booktitle}{\emph{Proceedings of the IEEE/CVF conference on computer vision and pattern recognition}}. \bibinfo{pages}{10684--10695}.
\newblock


\bibitem[Saharia et~al\mbox{.}(2022)]%
        {saharia2022photorealistic}
\bibfield{author}{\bibinfo{person}{Chitwan Saharia}, \bibinfo{person}{William Chan}, \bibinfo{person}{Saurabh Saxena}, \bibinfo{person}{Lala Li}, \bibinfo{person}{Jay Whang}, \bibinfo{person}{Emily~L Denton}, \bibinfo{person}{Kamyar Ghasemipour}, \bibinfo{person}{Raphael Gontijo~Lopes}, \bibinfo{person}{Burcu Karagol~Ayan}, \bibinfo{person}{Tim Salimans}, {et~al\mbox{.}}} \bibinfo{year}{2022}\natexlab{}.
\newblock \showarticletitle{Photorealistic text-to-image diffusion models with deep language understanding}.
\newblock \bibinfo{journal}{\emph{Advances in neural information processing systems}}  \bibinfo{volume}{35} (\bibinfo{year}{2022}), \bibinfo{pages}{36479--36494}.
\newblock


\bibitem[Shi et~al\mbox{.}(2023)]%
        {shi2023mvdream}
\bibfield{author}{\bibinfo{person}{Yichun Shi}, \bibinfo{person}{Peng Wang}, \bibinfo{person}{Jianglong Ye}, \bibinfo{person}{Mai Long}, \bibinfo{person}{Kejie Li}, {and} \bibinfo{person}{Xiao Yang}.} \bibinfo{year}{2023}\natexlab{}.
\newblock \showarticletitle{Mvdream: Multi-view diffusion for 3d generation}.
\newblock \bibinfo{journal}{\emph{arXiv preprint arXiv:2308.16512}} (\bibinfo{year}{2023}).
\newblock


\bibitem[Shum et~al\mbox{.}(2023)]%
        {shum2023conditional}
\bibfield{author}{\bibinfo{person}{Ka~Chun Shum}, \bibinfo{person}{Hong-Wing Pang}, \bibinfo{person}{Binh-Son Hua}, \bibinfo{person}{Duc~Thanh Nguyen}, {and} \bibinfo{person}{Sai-Kit Yeung}.} \bibinfo{year}{2023}\natexlab{}.
\newblock \showarticletitle{Conditional 360-degree image synthesis for immersive indoor scene decoration}. In \bibinfo{booktitle}{\emph{Proceedings of the IEEE/CVF International Conference on Computer Vision}}. \bibinfo{pages}{4478--4488}.
\newblock


\bibitem[Sohl-Dickstein et~al\mbox{.}(2015)]%
        {sohl2015deep}
\bibfield{author}{\bibinfo{person}{Jascha Sohl-Dickstein}, \bibinfo{person}{Eric Weiss}, \bibinfo{person}{Niru Maheswaranathan}, {and} \bibinfo{person}{Surya Ganguli}.} \bibinfo{year}{2015}\natexlab{}.
\newblock \showarticletitle{Deep unsupervised learning using nonequilibrium thermodynamics}. In \bibinfo{booktitle}{\emph{International conference on machine learning}}. pmlr, \bibinfo{pages}{2256--2265}.
\newblock


\bibitem[Song et~al\mbox{.}(2020)]%
        {song2020denoising}
\bibfield{author}{\bibinfo{person}{Jiaming Song}, \bibinfo{person}{Chenlin Meng}, {and} \bibinfo{person}{Stefano Ermon}.} \bibinfo{year}{2020}\natexlab{}.
\newblock \showarticletitle{Denoising diffusion implicit models}.
\newblock \bibinfo{journal}{\emph{arXiv preprint arXiv:2010.02502}} (\bibinfo{year}{2020}).
\newblock


\bibitem[Song and Ermon(2019)]%
        {song2019generative}
\bibfield{author}{\bibinfo{person}{Yang Song} {and} \bibinfo{person}{Stefano Ermon}.} \bibinfo{year}{2019}\natexlab{}.
\newblock \showarticletitle{Generative modeling by estimating gradients of the data distribution}.
\newblock \bibinfo{journal}{\emph{Advances in neural information processing systems}}  \bibinfo{volume}{32} (\bibinfo{year}{2019}).
\newblock


\bibitem[Tan et~al\mbox{.}(2024)]%
        {tan2024imagine360}
\bibfield{author}{\bibinfo{person}{Jing Tan}, \bibinfo{person}{Shuai Yang}, \bibinfo{person}{Tong Wu}, \bibinfo{person}{Jingwen He}, \bibinfo{person}{Yuwei Guo}, \bibinfo{person}{Ziwei Liu}, {and} \bibinfo{person}{Dahua Lin}.} \bibinfo{year}{2024}\natexlab{}.
\newblock \showarticletitle{Imagine360: Immersive 360 Video Generation from Perspective Anchor}.
\newblock \bibinfo{journal}{\emph{arXiv preprint arXiv:2412.03552}} (\bibinfo{year}{2024}).
\newblock


\bibitem[Tang et~al\mbox{.}(2023)]%
        {tang2023mvdiffusionenablingholisticmultiview}
\bibfield{author}{\bibinfo{person}{Shitao Tang}, \bibinfo{person}{Fuyang Zhang}, \bibinfo{person}{Jiacheng Chen}, \bibinfo{person}{Peng Wang}, {and} \bibinfo{person}{Yasutaka Furukawa}.} \bibinfo{year}{2023}\natexlab{}.
\newblock \bibinfo{title}{MVDiffusion: Enabling Holistic Multi-view Image Generation with Correspondence-Aware Diffusion}.
\newblock
\showeprint[arxiv]{2307.01097}~[cs.CV]
\urldef\tempurl%
\url{https://arxiv.org/abs/2307.01097}
\showURL{%
\tempurl}


\bibitem[Unterthiner et~al\mbox{.}(2018)]%
        {unterthiner2018towards}
\bibfield{author}{\bibinfo{person}{Thomas Unterthiner}, \bibinfo{person}{Sjoerd Van~Steenkiste}, \bibinfo{person}{Karol Kurach}, \bibinfo{person}{Raphael Marinier}, \bibinfo{person}{Marcin Michalski}, {and} \bibinfo{person}{Sylvain Gelly}.} \bibinfo{year}{2018}\natexlab{}.
\newblock \showarticletitle{Towards accurate generative models of video: A new metric \& challenges}.
\newblock \bibinfo{journal}{\emph{arXiv preprint arXiv:1812.01717}} (\bibinfo{year}{2018}).
\newblock


\bibitem[Voleti et~al\mbox{.}(2024)]%
        {voleti2024sv3d}
\bibfield{author}{\bibinfo{person}{Vikram Voleti}, \bibinfo{person}{Chun-Han Yao}, \bibinfo{person}{Mark Boss}, \bibinfo{person}{Adam Letts}, \bibinfo{person}{David Pankratz}, \bibinfo{person}{Dmitry Tochilkin}, \bibinfo{person}{Christian Laforte}, \bibinfo{person}{Robin Rombach}, {and} \bibinfo{person}{Varun Jampani}.} \bibinfo{year}{2024}\natexlab{}.
\newblock \showarticletitle{Sv3d: Novel multi-view synthesis and 3d generation from a single image using latent video diffusion}. In \bibinfo{booktitle}{\emph{European Conference on Computer Vision}}. Springer, \bibinfo{pages}{439--457}.
\newblock


\bibitem[Wang et~al\mbox{.}(2024c)]%
        {wang20244real}
\bibfield{author}{\bibinfo{person}{Chaoyang Wang}, \bibinfo{person}{Peiye Zhuang}, \bibinfo{person}{Tuan~Duc Ngo}, \bibinfo{person}{Willi Menapace}, \bibinfo{person}{Aliaksandr Siarohin}, \bibinfo{person}{Michael Vasilkovsky}, \bibinfo{person}{Ivan Skorokhodov}, \bibinfo{person}{Sergey Tulyakov}, \bibinfo{person}{Peter Wonka}, {and} \bibinfo{person}{Hsin-Ying Lee}.} \bibinfo{year}{2024}\natexlab{c}.
\newblock \showarticletitle{4Real-Video: Learning Generalizable Photo-Realistic 4D Video Diffusion}.
\newblock \bibinfo{journal}{\emph{arXiv preprint arXiv:2412.04462}} (\bibinfo{year}{2024}).
\newblock


\bibitem[Wang et~al\mbox{.}(2024b)]%
        {wang2024customizing}
\bibfield{author}{\bibinfo{person}{Hai Wang}, \bibinfo{person}{Xiaoyu Xiang}, \bibinfo{person}{Yuchen Fan}, {and} \bibinfo{person}{Jing-Hao Xue}.} \bibinfo{year}{2024}\natexlab{b}.
\newblock \showarticletitle{Customizing 360-degree panoramas through text-to-image diffusion models}. In \bibinfo{booktitle}{\emph{Proceedings of the IEEE/CVF Winter Conference on Applications of Computer Vision}}. \bibinfo{pages}{4933--4943}.
\newblock


\bibitem[Wang et~al\mbox{.}(2023a)]%
        {wang2023360}
\bibfield{author}{\bibinfo{person}{Jionghao Wang}, \bibinfo{person}{Ziyu Chen}, \bibinfo{person}{Jun Ling}, \bibinfo{person}{Rong Xie}, {and} \bibinfo{person}{Li Song}.} \bibinfo{year}{2023}\natexlab{a}.
\newblock \showarticletitle{360-degree panorama generation from few unregistered nfov images}.
\newblock \bibinfo{journal}{\emph{arXiv preprint arXiv:2308.14686}} (\bibinfo{year}{2023}).
\newblock


\bibitem[Wang et~al\mbox{.}(2024a)]%
        {wang2024360dvd}
\bibfield{author}{\bibinfo{person}{Qian Wang}, \bibinfo{person}{Weiqi Li}, \bibinfo{person}{Chong Mou}, \bibinfo{person}{Xinhua Cheng}, {and} \bibinfo{person}{Jian Zhang}.} \bibinfo{year}{2024}\natexlab{a}.
\newblock \showarticletitle{360dvd: Controllable panorama video generation with 360-degree video diffusion model}. In \bibinfo{booktitle}{\emph{Proceedings of the IEEE/CVF Conference on Computer Vision and Pattern Recognition}}. \bibinfo{pages}{6913--6923}.
\newblock


\bibitem[Wang et~al\mbox{.}(2023b)]%
        {wang2023cogvlm}
\bibfield{author}{\bibinfo{person}{Weihan Wang}, \bibinfo{person}{Qingsong Lv}, \bibinfo{person}{Wenmeng Yu}, \bibinfo{person}{Wenyi Hong}, \bibinfo{person}{Ji Qi}, \bibinfo{person}{Yan Wang}, \bibinfo{person}{Junhui Ji}, \bibinfo{person}{Zhuoyi Yang}, \bibinfo{person}{Lei Zhao}, \bibinfo{person}{Xixuan Song}, \bibinfo{person}{Jiazheng Xu}, \bibinfo{person}{Bin Xu}, \bibinfo{person}{Juanzi Li}, \bibinfo{person}{Yuxiao Dong}, \bibinfo{person}{Ming Ding}, {and} \bibinfo{person}{Jie Tang}.} \bibinfo{year}{2023}\natexlab{b}.
\newblock \bibinfo{title}{CogVLM: Visual Expert for Pretrained Language Models}.
\newblock
\showeprint[arxiv]{2311.03079}~[cs.CV]


\bibitem[Wu et~al\mbox{.}(2023a)]%
        {wu2023q}
\bibfield{author}{\bibinfo{person}{Haoning Wu}, \bibinfo{person}{Zicheng Zhang}, \bibinfo{person}{Weixia Zhang}, \bibinfo{person}{Chaofeng Chen}, \bibinfo{person}{Liang Liao}, \bibinfo{person}{Chunyi Li}, \bibinfo{person}{Yixuan Gao}, \bibinfo{person}{Annan Wang}, \bibinfo{person}{Erli Zhang}, \bibinfo{person}{Wenxiu Sun}, {et~al\mbox{.}}} \bibinfo{year}{2023}\natexlab{a}.
\newblock \showarticletitle{Q-align: Teaching lmms for visual scoring via discrete text-defined levels}.
\newblock \bibinfo{journal}{\emph{arXiv preprint arXiv:2312.17090}} (\bibinfo{year}{2023}).
\newblock


\bibitem[Wu et~al\mbox{.}(2024)]%
        {wu2024cat4d}
\bibfield{author}{\bibinfo{person}{Rundi Wu}, \bibinfo{person}{Ruiqi Gao}, \bibinfo{person}{Ben Poole}, \bibinfo{person}{Alex Trevithick}, \bibinfo{person}{Changxi Zheng}, \bibinfo{person}{Jonathan~T Barron}, {and} \bibinfo{person}{Aleksander Holynski}.} \bibinfo{year}{2024}\natexlab{}.
\newblock \showarticletitle{Cat4d: Create anything in 4d with multi-view video diffusion models}.
\newblock \bibinfo{journal}{\emph{arXiv preprint arXiv:2411.18613}} (\bibinfo{year}{2024}).
\newblock


\bibitem[Wu et~al\mbox{.}(2023b)]%
        {wu2023ipo}
\bibfield{author}{\bibinfo{person}{Tianhao Wu}, \bibinfo{person}{Chuanxia Zheng}, {and} \bibinfo{person}{Tat-Jen Cham}.} \bibinfo{year}{2023}\natexlab{b}.
\newblock \showarticletitle{Ipo-ldm: Depth-aided 360-degree indoor rgb panorama outpainting via latent diffusion model}.
\newblock \bibinfo{journal}{\emph{arXiv preprint arXiv:2307.03177}}  \bibinfo{volume}{3} (\bibinfo{year}{2023}).
\newblock


\bibitem[Xu et~al\mbox{.}(2024)]%
        {xu2024comp4d}
\bibfield{author}{\bibinfo{person}{Dejia Xu}, \bibinfo{person}{Hanwen Liang}, \bibinfo{person}{Neel~P Bhatt}, \bibinfo{person}{Hezhen Hu}, \bibinfo{person}{Hanxue Liang}, \bibinfo{person}{Konstantinos~N Plataniotis}, {and} \bibinfo{person}{Zhangyang Wang}.} \bibinfo{year}{2024}\natexlab{}.
\newblock \showarticletitle{Comp4d: Llm-guided compositional 4d scene generation}.
\newblock \bibinfo{journal}{\emph{arXiv preprint arXiv:2403.16993}} (\bibinfo{year}{2024}).
\newblock


\bibitem[Yang et~al\mbox{.}(2024)]%
        {yang2024layerpano3d}
\bibfield{author}{\bibinfo{person}{Shuai Yang}, \bibinfo{person}{Jing Tan}, \bibinfo{person}{Mengchen Zhang}, \bibinfo{person}{Tong Wu}, \bibinfo{person}{Yixuan Li}, \bibinfo{person}{Gordon Wetzstein}, \bibinfo{person}{Ziwei Liu}, {and} \bibinfo{person}{Dahua Lin}.} \bibinfo{year}{2024}\natexlab{}.
\newblock \showarticletitle{Layerpano3d: Layered 3d panorama for hyper-immersive scene generation}.
\newblock \bibinfo{journal}{\emph{arXiv preprint arXiv:2408.13252}} (\bibinfo{year}{2024}).
\newblock


\bibitem[Yin et~al\mbox{.}(2023)]%
        {yin20234dgen}
\bibfield{author}{\bibinfo{person}{Yuyang Yin}, \bibinfo{person}{Dejia Xu}, \bibinfo{person}{Zhangyang Wang}, \bibinfo{person}{Yao Zhao}, {and} \bibinfo{person}{Yunchao Wei}.} \bibinfo{year}{2023}\natexlab{}.
\newblock \showarticletitle{4dgen: Grounded 4d content generation with spatial-temporal consistency}.
\newblock \bibinfo{journal}{\emph{arXiv preprint arXiv:2312.17225}} (\bibinfo{year}{2023}).
\newblock


\bibitem[Yu et~al\mbox{.}(2024)]%
        {yu2024wonderworld}
\bibfield{author}{\bibinfo{person}{Hong-Xing Yu}, \bibinfo{person}{Haoyi Duan}, \bibinfo{person}{Charles Herrmann}, \bibinfo{person}{William~T Freeman}, {and} \bibinfo{person}{Jiajun Wu}.} \bibinfo{year}{2024}\natexlab{}.
\newblock \showarticletitle{Wonderworld: Interactive 3d scene generation from a single image}.
\newblock \bibinfo{journal}{\emph{arXiv preprint arXiv:2406.09394}} (\bibinfo{year}{2024}).
\newblock


\bibitem[Zhang et~al\mbox{.}(2024b)]%
        {zhang2024taming}
\bibfield{author}{\bibinfo{person}{Cheng Zhang}, \bibinfo{person}{Qianyi Wu}, \bibinfo{person}{Camilo~Cruz Gambardella}, \bibinfo{person}{Xiaoshui Huang}, \bibinfo{person}{Dinh Phung}, \bibinfo{person}{Wanli Ouyang}, {and} \bibinfo{person}{Jianfei Cai}.} \bibinfo{year}{2024}\natexlab{b}.
\newblock \showarticletitle{Taming stable diffusion for text to 360 panorama image generation}. In \bibinfo{booktitle}{\emph{Proceedings of the IEEE/CVF Conference on Computer Vision and Pattern Recognition}}. \bibinfo{pages}{6347--6357}.
\newblock


\bibitem[Zhang et~al\mbox{.}(2024a)]%
        {zhang2024monst3r}
\bibfield{author}{\bibinfo{person}{Junyi Zhang}, \bibinfo{person}{Charles Herrmann}, \bibinfo{person}{Junhwa Hur}, \bibinfo{person}{Varun Jampani}, \bibinfo{person}{Trevor Darrell}, \bibinfo{person}{Forrester Cole}, \bibinfo{person}{Deqing Sun}, {and} \bibinfo{person}{Ming-Hsuan Yang}.} \bibinfo{year}{2024}\natexlab{a}.
\newblock \showarticletitle{Monst3r: A simple approach for estimating geometry in the presence of motion}.
\newblock \bibinfo{journal}{\emph{arXiv preprint arXiv:2410.03825}} (\bibinfo{year}{2024}).
\newblock


\bibitem[Zhou et~al\mbox{.}(2024)]%
        {zhou2024dreamscene360}
\bibfield{author}{\bibinfo{person}{Shijie Zhou}, \bibinfo{person}{Zhiwen Fan}, \bibinfo{person}{Dejia Xu}, \bibinfo{person}{Haoran Chang}, \bibinfo{person}{Pradyumna Chari}, \bibinfo{person}{Tejas Bharadwaj}, \bibinfo{person}{Suya You}, \bibinfo{person}{Zhangyang Wang}, {and} \bibinfo{person}{Achuta Kadambi}.} \bibinfo{year}{2024}\natexlab{}.
\newblock \showarticletitle{Dreamscene360: Unconstrained text-to-3d scene generation with panoramic gaussian splatting}. In \bibinfo{booktitle}{\emph{European Conference on Computer Vision}}. Springer, \bibinfo{pages}{324--342}.
\newblock


\bibitem[Zhuang et~al\mbox{.}(2022)]%
        {zhuang2022acdnet}
\bibfield{author}{\bibinfo{person}{Chuanqing Zhuang}, \bibinfo{person}{Zhengda Lu}, \bibinfo{person}{Yiqun Wang}, \bibinfo{person}{Jun Xiao}, {and} \bibinfo{person}{Ying Wang}.} \bibinfo{year}{2022}\natexlab{}.
\newblock \showarticletitle{ACDNet: Adaptively combined dilated convolution for monocular panorama depth estimation}. In \bibinfo{booktitle}{\emph{Proceedings of the AAAI conference on artificial intelligence}}, Vol.~\bibinfo{volume}{36}. \bibinfo{pages}{3653--3661}.
\newblock


\bibitem[Zuo et~al\mbox{.}(2024)]%
        {zuo2024videomv}
\bibfield{author}{\bibinfo{person}{Qi Zuo}, \bibinfo{person}{Xiaodong Gu}, \bibinfo{person}{Lingteng Qiu}, \bibinfo{person}{Yuan Dong}, \bibinfo{person}{Weihao Yuan}, \bibinfo{person}{Rui Peng}, \bibinfo{person}{Siyu Zhu}, \bibinfo{person}{Liefeng Bo}, \bibinfo{person}{Zilong Dong}, \bibinfo{person}{Qixing Huang}, {et~al\mbox{.}}} \bibinfo{year}{2024}\natexlab{}.
\newblock \showarticletitle{Videomv: Consistent multi-view generation based on large video generative model}.
\newblock  (\bibinfo{year}{2024}).
\newblock


\end{thebibliography}

\clearpage
%
\appendix

\section{Implementation Details}

\subsection{Dual-branch Video Generation Model}
\textbf{Dataset.} We use the Matterport3D dataset~\cite{chang2017matterport3d}, which contains 10,800 panoramic images from 90 building-scale indoor scenes, for training the dual-branch image diffusion model. Text descriptions for both panoramic and perspective views are obtained using BLIP-2~\cite{li2023blip} following~\cite{zhang2024taming}.
Also, we adopt WEB360~\cite{wang2024360dvd}, a high-quality panorama video dataset comprising 2,114 text-video pairs in 720P ERP format, with each video consisting of 100 frames. We uniformly sample 10 frames per video for image-level training.
WEB360 is used for finetuning the dual-branch video model.
Captions for perspective images and videos are generated using CogVLM~\cite{wang2023cogvlm}.

\textbf{Training Details.} We use Stable Diffusion 1.5~\cite{blattmann2023stable} as the backbone for both branches, with rank-4 LoRA modules added to the spatial layers. Following~\cite{zhang2024taming,shi2023mvdream}, we train the image diffusion model for 10 epochs using the AdamW optimizer with a batch size of 64 and a learning rate of 2e-4.
After pretraining, we integrate the motion module from AnimateDiff~\cite{guo2023animatediff} and bidirectional cross-attention modules, and train the video diffusion model on WEB360 for 10K steps using AdamW (batch size 8, learning rate 1e-4). We adopt DDIM sampling with 50 steps for inference. All training is performed on 8 NVIDIA A100 GPUs.

\subsection{Geometry-aligned Reconstruction Model}
\textbf{Training Details.}
In 3DGS optimization, each panoramic frame is projected into $K = 20$ overlapping tangent perspective views to supervise Gaussian Splatting training.
During projection, the camera's orientation is randomly initialized by uniformly sampling Euler angles across all three rotational axes (X, Y, Z) over the full 360-degree sphere.
The camera's field of view (FOV) is set to 90-degree with perspective resolution fixed at 512 × 512 pixels. 
We extract extrinsic parameters from the estimated camera poses $\{{P}^{t}\}_{t=1}^T$ and use the intrinsic parameters to initialize scene cameras.
Each panoramic frame is optimized for 15,000 epochs, employing both RGB loss and geometric correspondence loss throughout the entire training process. Additionally, the distillation semantic loss is applied between epochs 5,400 and 9,000.
We set losses weights as $\lambda_{1}=0.8, \lambda_{ssim}=0.2, \lambda_{lpips}=0.05, \lambda_{sem}=1, \lambda_{geo}=0.05$.
All training and experiments are performed on 4 NVIDIA 3090 GPUs.

\begin{figure*}[ht]
    \centering
    \includegraphics[width=\textwidth]{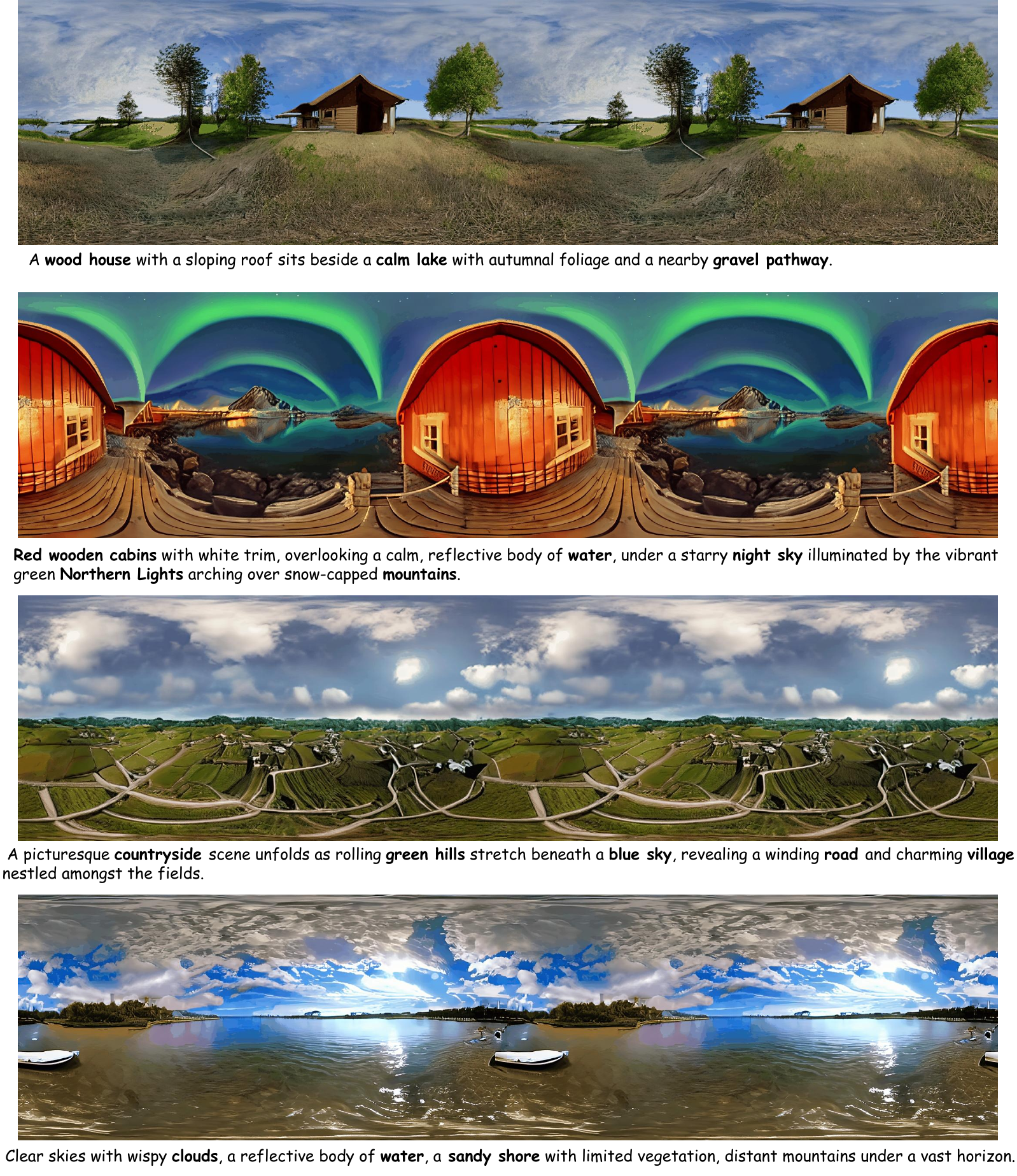}
    \caption{Loop Consistency Visualization.}
    \label{fig:connect1}
\end{figure*}

\begin{figure*}[ht]
    \centering
    \includegraphics[width=\textwidth]{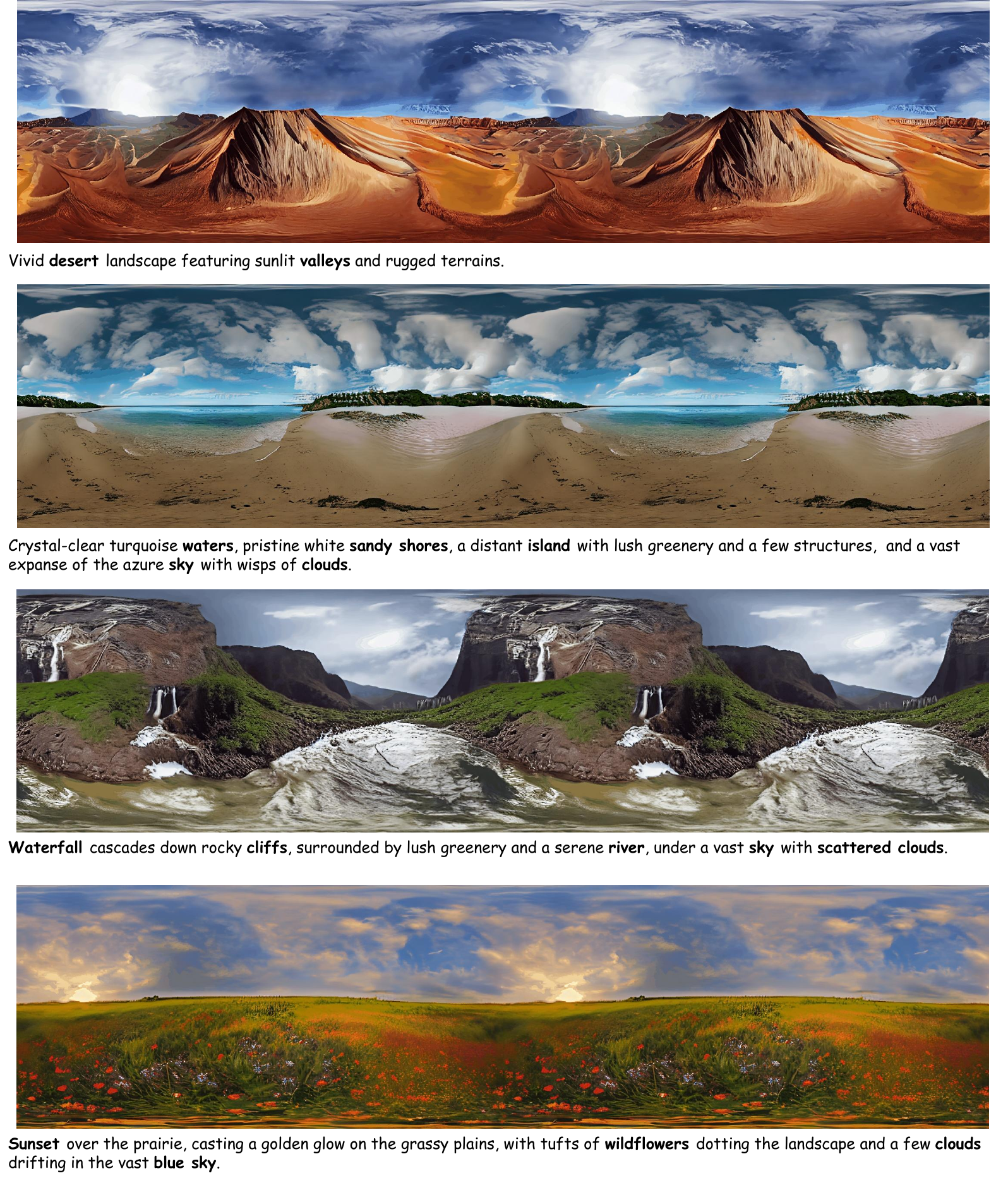}
    \caption{Loop Consistency Visualization.}
    \label{fig:connect2}
\end{figure*}

\section{Loop Consistency Visualization}
We stitch the left and right ends of the generated panoramic videos and visualize the results in Fig.~\ref{fig:connect1} and Fig.~\ref{fig:connect2}. As shown, our method produces panoramas with smooth and seamless transitions at the boundaries, demonstrating strong consistency and continuity across the entire 360-degree view. \textbf{We provide more video demos in the supplementary materials.}

\section{Limitations and Future Work}
In this work, we propose a novel text-to-dynamic panoramic 4D generation pipeline that integrates a dual-branch video generation model with a geometry-consistent reconstruction framework. While our method demonstrates strong performance across various benchmarks, there are still some limitations. First, our model is built upon pretrained image diffusion backbones, such as Stable Diffusion, to leverage their powerful generative capabilities. However, this reliance also imposes constraints, as the performance and flexibility of our approach are inherently limited by the capabilities and biases of the underlying pretrained models. Second, the scarcity of high-quality and diverse panoramic video datasets poses a challenge for generalization. While we make use of WEB360 and Matterport3D, these datasets still lack the variety and coverage to fully support broad, open-world scene generation and reconstruction. 
In future work, we plan to explore more robust and expressive base models to further enhance generation quality and flexibility. Also, we aim to curate or construct larger-scale, high-diversity panoramic video datasets to improve the generalization and adaptability of our pipeline.

\end{document}